\documentclass[a4paper,fleqn]{cas-dc}

\usepackage[numbers]{natbib}
\usepackage{amsmath}

\def\tsc#1{\csdef{#1}{\textsc{\lowercase{#1}}\xspace}}
\tsc{WGM}
\tsc{QE}
\tsc{EP}
\tsc{PMS}
\tsc{BEC}
\tsc{DE}

\begin{document}
\let\WriteBookmarks\relax
\def\floatpagepagefraction{1}
\def\textpagefraction{.001}
\shorttitle{Backbones-Review: Feature Extraction Networks for Deep Learning and Deep Reinforcement Learning Approaches}
\shortauthors{Elharrouss et~al.}

\title [mode = title]{Backbones-Review: Feature Extraction Networks for Deep Learning and Deep Reinforcement Learning Approaches}

\author[1]{Omar Elharrouss}[]
\cormark[1]
\ead{Omar Elharrouss}

\author[1]{Younes Akbari}[]
\ead{akbari\_younes@yahoo.com}

\author[1]{Noor Almaadeed}[]
\ead{n.alali@qu.edu.qa}

\author[1]{Somaya Al-Maadeed}[]
\ead{S\_alali@qu.edu.qa}

\address[1]{Department of Computer Science and Engineering, Qatar University,Doha, Qatar}

\cortext[cor1]{Corresponding author}
\cortext[cor2]{Principal corresponding author}

\begin{abstract}
To understand the real world using various types of data, Artificial Intelligence (AI) is the most used technique nowadays. While finding the pattern within the analyzed data represents the main task. This is performed by extracting representative features step, which is proceeded using the statistical algorithms or using some specific filters. However, the selection of useful features from large-scale data represented a crucial challenge. Now, with the development of convolution neural networks (CNNs), the feature extraction operation has become more automatic and easier. CNNs allow to work on large-scale size of data, as well as cover different scenarios for a specific task. For computer vision tasks, convolutional networks are used to extract features also for the other parts of a deep learning model. The selection of a suitable network for feature extraction or the other parts of a DL model is not random work. So, the implementation of such a model can be related to the target task as well as the computational complexity of it. Many networks have been proposed and become the famous networks used for any DL models in any AI task. These networks are exploited for feature extraction or at the beginning of any DL model which is named backbones. A backbone is a known network trained in many other tasks before and demonstrates its effectiveness. In this paper, an overview of the existing backbones, e.g. VGGs, ResNets, DenseNet, etc, is given with a detailed description. Also, a couple of computer vision tasks are discussed by providing a review of each task regarding the backbones used. In addition, a comparison in terms of performance is also provided, based on the backbone used for each task.
\end{abstract}





\maketitle

\section{Introduction}
Artificial intelligence is one of the most research topics dedicated to understand the real world from various types of data. For time series data, the classification, recognition of patterns, and then make decisions become easier using AI techniques \cite{i1}. On image/video data, the processing purpose is to detect or recognize an object, classify the behaviors of a person, analyze and understand a monitored scene, or prevent abnormal actions in an event, etc. This is performed using different features which are used to improve the performance of each task  \cite{i2}. These feature are exploited by various techniques starting from traditional statistical methods, passing by neural networks and deep learning, to deep reinforcement learning.
 
Before, Statistical-based methods were suffer from the finding of patterns, due to the different scenarios of a task, the variation of different aspects, and the data content that should be analyzed. With the Machine learning (ML) techniques most of these challenges still exist but the learning from various features lead to an improvement in terms of performance  \cite{i3}. Due to the large-scale datasets that need to be analyzed, ML algorithms can't process these among of data which prohibit the analyzing of a set of situations that can be contained in it. With the introduction of deep learning techniques including convolutional neural networks (CNNs), the processing of large-scale datasets becomes doable. Also the automatic learning from large set leads to a good learning from a large-scale of features. Nowadays, the use of deep learning models and the combination of deep learning and Reinforcement learning (RL) techniques \cite{i4}, which known under the name of Deep Reinforcement Learning (DRL), lead to a real improvement. While Reinforcement learning (RL) is one of modern machine learning technologies in which learning is carried out through interaction with the environment and allows taking into account results of decisions and further actions based on solutions of corresponding tasks \cite{X}. With DL and DRL, data analysis overcome many challenges with the possibility to learn from large-scale datasets as well as the processing of different scenarios which can lead to an generic learning, then to robust methods that can work in different environments \cite{AlexNet} . For computer vision tasks, features are extracted using different convolutional networks (backbones), while the processing of images/videos can be reached with a multi-task learning that can handle the variation of scales, positions of objects, the low resolutions, etc.

In the literature, there is luck of papers that compared the proposed features extraction networks for deep-learning-based techniques \cite{X,VGG} . For computer vision tasks , the choose of suitable network (Backbone) for features extraction can be costly, due to the fact that some tasks are used some specific backbones while its not suitable for others  \cite{o15,o16}. In this paper, we attempted to collect and describe various existing backbones used for features extraction. Presenting the specific backbones used for each task is provided also. For that, a set of backbones are described in this paper including AlexNet, GoogleNet, VGGs, ResNet, Inceptions, Xception, DenseNet, Inception-ResNet, ResNeXt, SqueezeNet, MobileNet, EfficientNet and many other. Some of computer visions tasks that used these backbones for features extraction has been also discussed such as image classification, object detection, face recognition, panoptic segmentation, action recognition, ect.  
 Accordingly, this paper presents a set of contributions that can be summarized as follows:
\begin{itemize}
    \item Presentation of various networks used as backbone for deep learning (DL) and  Deep Reinforcement Learning (DRL) techniques.
    \item An overview of some computer vision tasks based on the backbones used. 
    \item Evaluation and comparison study of each task based on the backbone used. 
    \item Summarization of the most backbones used for each tasks.
    \item Presentation of deep learning challenges as well as some futures directions .
\end{itemize}

 The remainder of this paper is organized as follows. Overview of the existing backbones is presented in Section 2. The computer vision tasks that used these backbones is presented in Section 3. Comparison and discussion of each task according to backbone used  in Section 4. Challenges and future directions are  provided in Section 5.The conclusion is provided in Section 6. 
 
\begin{figure}[t!]
  \centering
  \footnotesize
  \begin{tabular}[b]{c}
    \includegraphics[width=.37\linewidth]{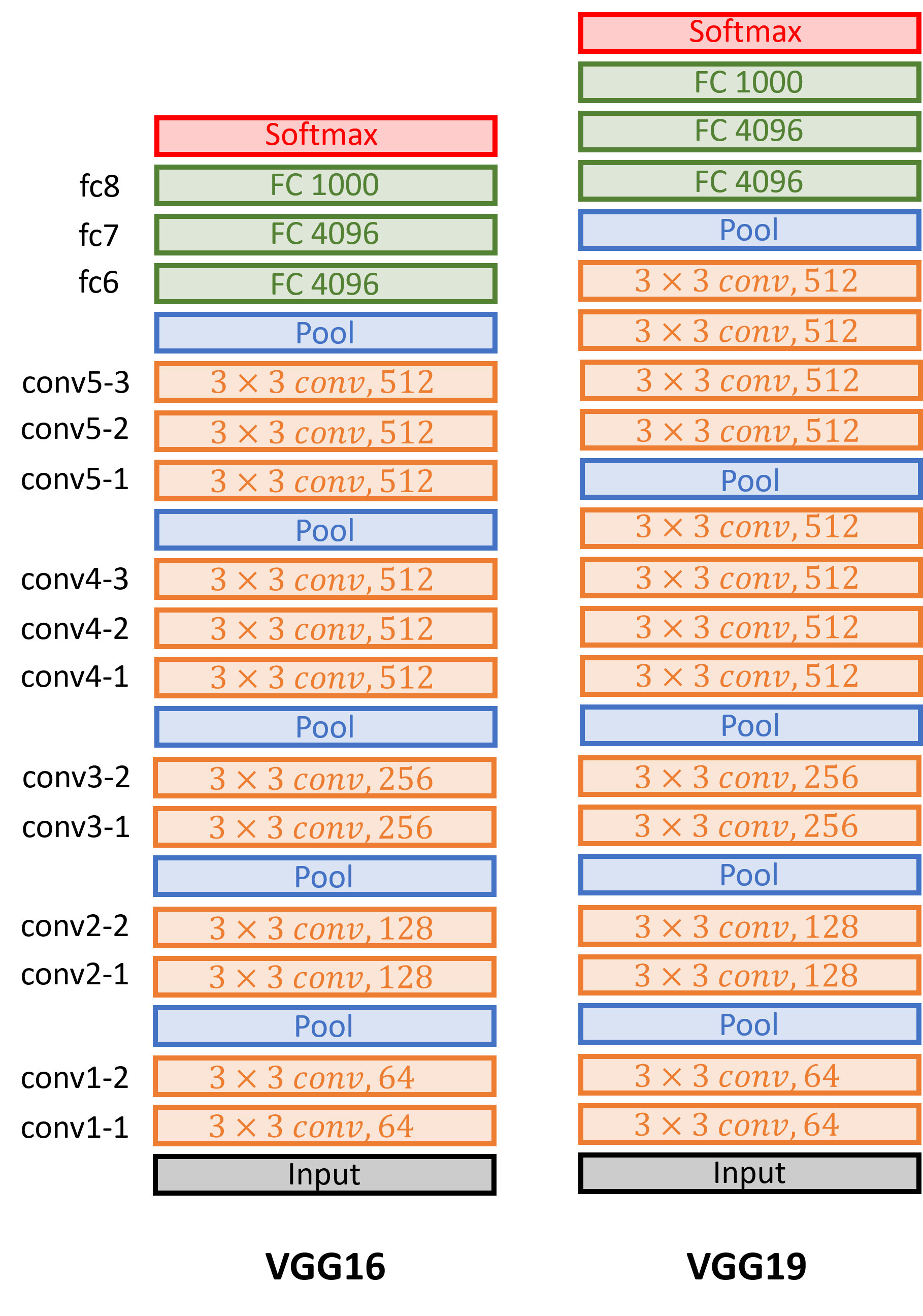} \\
    \textbf{(a) VGG }
  \end{tabular} 
    \begin{tabular}[b]{c}
    \includegraphics[width=.37\linewidth]{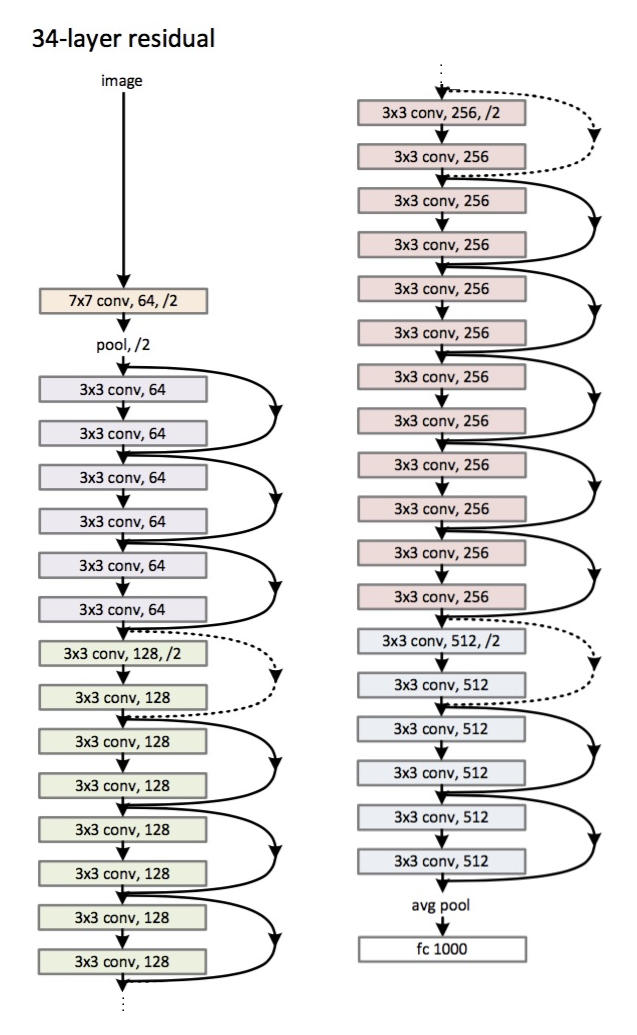} \\
    \textbf{(b) ResNet}
  \end{tabular} 
 
  \caption{ VGG and ResNet architectures.}
  \label{VGGback}
\end{figure}
\section{Backbone families}
Features extraction is the main step in data analysis domains. Before the features extraction was provided using statistical algorithms or some filter applied to the input data to be used in the next processing steps. With the introduction of machine and deep learning (DL) techniques the use of neural networks make a revolutional evolution in term of performance and the the number of data that can be processed \cite{i1}. Then, the development of convolution neural networks (CNNs) make the work on large-scale size of data possible, also used for features extraction. The selection of a CNN network for features extraction or the other part of a DL model, is not a random work  \cite{i2}. So, the implementation of such model can be related to the target task as well as the complexity of it. Some proposed networks becomes the famous networks used for different data analysis domains. These networks are used now for features extraction or in the beginning of any DL model and its named backbones. A backbone is the recognized architecture or network used for feature extraction and its trained in many other task before and demonstrate its effectiveness. In this section, a detailed description for each backbone used in deep learning models is provided.

\subsection{ AlexNet}
AlexNet is a CNN architecture developed by Krizhevsky et al. \cite{AlexNet} in 2012 for image classification. The model consist of a set of convolutional and max-pooling layers ended by 3 fully connected layers. The proposed networks contains 5 convolutions layers which make it a simple network. In addition, it trained using Rectified Linear Units (ReLUs) as activation. While the regularization function or Dropout is introduced to reduce overfitting in the fully-connected layers. Dropout proved it effectiveness for AlexNet and also for the deep learning model after. AlexNet has 60 million parameters and 650,000 neurons. This network is used for image classification on ImageNet dataset. Also, is used a backbone for many object detection and segmentation models such as R-CNN \cite{7} and HyperNet \cite{8}.

\subsection{VGGs}
The VGG family, that includes VGG-16 and VGG-19 \cite{VGG}, is one of the famous backbone used for computer vision and computer sciences tasks. The VGGs architecture are proved it effectiveness in many tasks including image classification and object detection, and many other tasks. While it widely used for many other architectures as backbone (for features extraction) for many other recognized models like R-CNN \cite{fast},Faster R-CNN \cite{faster}, and SSD \cite{ssd}.

For VGG-16 \cite{VGG}\footnote{https://github.com/pytorch/vision/blob/master/torchvision/models/vgg.py} in one of the fundamental deep learning backbones developed in 2014. VGG-16 contains 16 layers with 13 convolutional layers and 5 max-pooling layers and 3 fully connected layers. In addition, ReLU is used as activation. Comparing to AlexNet architecture, VGG has 8 more layers.  It has 138 million parameters. 

For VGG-19 \footnote{https://github.com/pytorch/vision/blob/master/torchvision/models/vgg.py} is a deeper version of VGG-16. It contains 3 more layers with 16 convolutional layers, 5 max-pooling layers and 3 fully connected layers.VGG-19 architecture has 144 million parameters. The architectures of VGG-16 and VGG-19 are are presented in Figure \ref{VGGback}.


\begin{figure}[t!]
  \centering
  \footnotesize
    \begin{tabular}[b]{c}
    \includegraphics[width=1\linewidth]{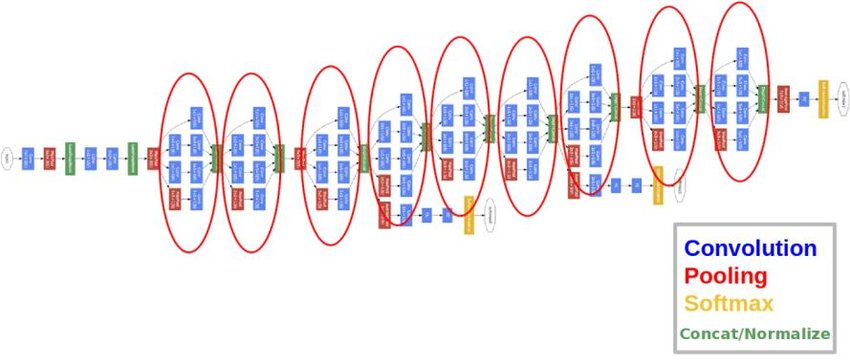}\\
    \textbf{(a) GoogleNet}
  \end{tabular}
  \begin{tabular}[b]{c}
    \includegraphics[width=1\linewidth]{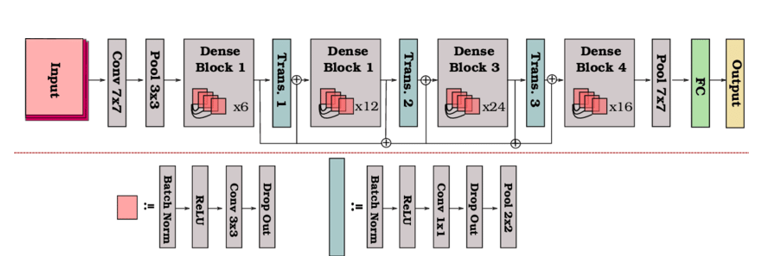}\\
     \textbf{(b) DenseNet}
  \end{tabular}

  \caption{ GoogleNet and DenseNet architectures}
  \label{figure:googlenet}
\end{figure}
\subsection{ResNets}
Unlike the previous CNN architecture, Residual Neural Network (ResNet)  \cite{ResNet}\footnote{https://github.com/pytorch/vision/blob/master/torchvision/models/resnet.py} is a CNN-based model while the residual networks is introduced. ResNet or Residual neural networks consists of some skip-connections or recurrent units between blocks of convolutional and pooling layers. Also, the block are followed by a batch normalization \cite{17}. Like VGG family, ResNet has many versions including ResNet-34 and ResNet-50 with 26M parameters, ResNet-101 with 44M and ResNet-152 which is deeper with 152 layers like presented in Table \ref{tablepara}. ResNet-50 and ResNet-101 are used widely for object detection and semantic segmentation.  ResNet also used for other deep learning architectures like Faster R-CNN \cite{faster} and  R-FCN \cite{20}, etc.

\subsection{Inception-v1 (GoogleNet)}
Inception-V1 or GoogleNet\footnote{https://github.com/Lornatang/GoogLeNet-PyTorch} is one of the most used convolutional neural networks as backbone for many computer science applications \cite{Inceptionv1} . It developed based on blocks of inception. Each block is a set of convolution layers, while the filters used vary from 1x1, 3x3 to 5x5, which allow a multi-scale learning. The size of each filter make the a variation of dimension between blocks. Also GoogleNet uses global average pooling instead of Max-pooling used in AlexNet and VGG. Inception-v1 architecture is illustrated in Figure \ref{figure:googlenet}.

\subsection{DenseNet}
In traditional CNNs the number of layers $L$ is the same as the number of connections. While the connection between layers can have an impact on the learning. For that the authors in \cite{DenseNet} \footnote{https://github.com/liuzhuang13/DenseNet}, introduced a new convolutional neural network architecture named DenseNet with $L(L+1)/2$ connections. For each layer, the outputs (feature maps) of all previous layers are used as input to the next layer. The network could work with very small output channel depths (ie. 12 filters per layer), which reduce the number of parameters. The number of filter used in each convolutional layer is initialized, and after each layer they used more filter than the previous layers with a constant of $k$ or named "growth rate". This make the number of parameters depend to $k$. Many versions of DenseNet have been proposed with different number of layers, such as DenseNet-121, DenseNet-169, DenseNet-201, DenseNet-264. The network has an image input size of $224\times224$.

\begin{figure}[t!]
  \centering
  \footnotesize

   \begin{tabular}[b]{c}
    \includegraphics[width=0.8\linewidth]{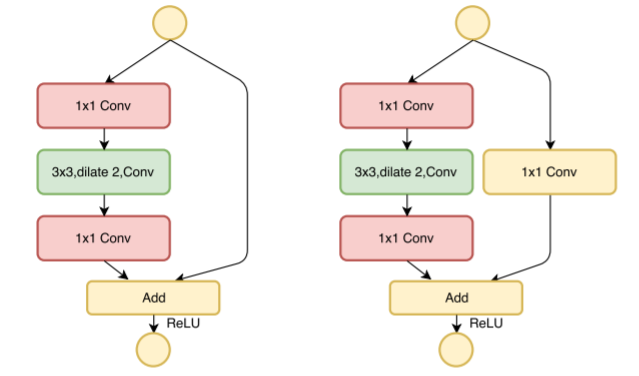} \\
    \textbf{(a) DetNet}
  \end{tabular}  
  \begin{tabular}[b]{c}
    \includegraphics[width=0.8\linewidth]{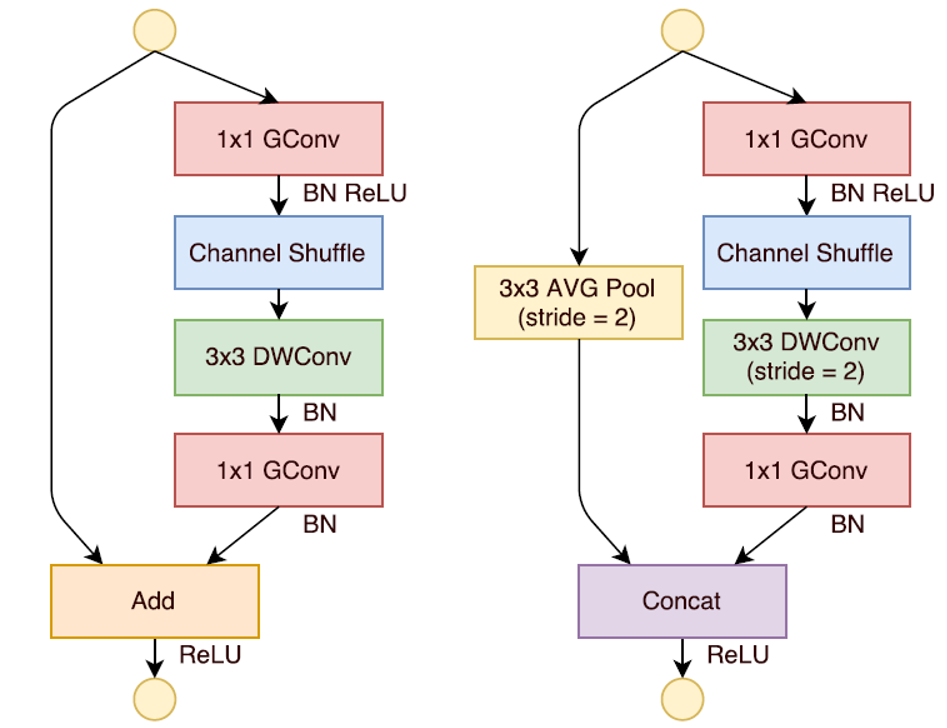} \\
    \textbf{(b) ShuffleNet}
  \end{tabular} 
  
  \caption{ DetNet and ShuffleNet architectures }
  \label{DetNet}
\end{figure}

\subsection{BN-Inception and Inception-v3}

The computational cost is the most challenge in any deep learning model. The changes of the parameters between the consecutive layers and the initialization of it as well as the selection of required learning rates, during the training, make the training slow. The researchers attempted to solve this by normalizing input layers, and apply this normalization in all network blocks. This operation is named by Batch Normalization (BN) which minimizes the impact of parameters initialization and permits the use of higher learning rates \cite{BNInception}. BN is applied to the Inception network (BN-Inception\footnote{https://github.com/Cadene/pretrained-models.pytorch/blob/master/pretrainedmodels/models/bninception.py}) and tested on ImageNet for image classification \cite{BNInception}. The obtained results are close to the Inception results on the same dataset with a lower computational time.  

\subsection{Inception-v2-v3}

The development of deep convolutional networks for the variety of computer vision tasks is often prohibited by the challenges of computational cost caused by the size of these models. For that, the implementation of networks with low number parameters is a real need especially to be able to use them in machines of low performance like mobiles and  Raspberry-Pis. In order to implement an efficient network with a low number of parameters, also using the architecture of Inception-v1 the authors in \cite{Inceptionv3} developed Inception-v2 \footnote{https://github.com/weiaicunzai/pytorch-cifar100/blob/master/models/inceptionv3.py} and Inception-v3 networks by replacing nn convolutional kernels in Inception-v1 by $3\times3$ convolutional kernels as well as using  $1\times1$ convolutional kernels with a proposed concatenation method. This strategy used less than 25 million parameters. The two networks are trained and tested on image classification dataset like ImageNet and CIFAR-100.

\begin{table*}
\begin{center}
\caption{Summarization of crowd counting methods}
\label{tablepara}
\begin{tabular}{|l|c|c|c|}
\hline
\textbf{Backbone} &	\textbf{Year } & \textbf{\# of parameters} &	\textbf{trained task} \\
\hline
AlexNet  \cite{AlexNet} & 2012& 60M 	 & Img-class  \\\hline
VGG-16  \cite{VGG}& 2014 & 	138M 	 &Img-class \\\hline
VGG-19  \cite{VGG}& 2014 & 	 144M 	 &  Img-class \\\hline
Inception-V1 (GoogleNet)  \cite{Inceptionv1}&2014 &  5 M & Img-class \\\hline

ResNet-18  \cite{ResNet} & 2015&  11.7 M &  Img-class  \\\hline
ResNet-34  \cite{ResNet} & 2015&  25.6 M &  Img-class \\\hline
ResNet-50   \cite{ResNet} & 2015&  26 M &  Img-class\\\hline
ResNet-101   \cite{ResNet} & 2015&  44.6 M &  Img-class \\\hline
ResNet-152   \cite{ResNet} & 2015&  230M & Img-class  \\\hline

Inception-v2 \cite{Inceptionv3}& 2015& 21.8M& Img-class  \\\hline
Inception-v3 \cite{Inceptionv3}& 2015&  21.8M & Img-class  \\\hline
Inception-ResNet-V2 \cite{Inception-ResNet} & 2015& 55 M & Img-class, obj-det \\\hline

Darknet-19 2015  \cite{DarkNet} & 2015&  20.8 M & Obj-det  \\\hline
Xception \cite{Xception}& 2017 & 22.9 M &  Img-class \\\hline

SqueezeNet 2016 \cite{SqueezeNet}& 2016&  1.24M & Img-class\\\hline
ShuffleNet\cite{shuffleNet}(g = 1) & 2017 & 143M M &Img-class, obj-det \\\hline
ShuffleNet-v2\cite{shuffleNetv2}(g = 1) & 2018 &  2.3 M &Img-class, obj-det  \\\hline
DenseNet-100 (k = 12)  \cite{DenseNet}& 2018&7.0M & Img-class \\\hline
DenseNet-100 (k = 24)  \cite{DenseNet}& 2018 &27.2M & Img-class \\\hline
DenseNet-250 (k = 24)  \cite{DenseNet}& 2018&15.3M & Img-class  \\\hline
DenseNet-190 (k = 40)  \cite{DenseNet}& 2018&25.6M & Img-class  \\\hline
DetNet \cite{Detnet}& 2018 &- & Img-class, obj-det \\\hline

EfficientNet B0 to B7 \cite{efficientnet}& 2020&  5.3 M,  to 66M & Img-class, obj-det \\\hline 
MobileNet \cite{Mobilenet} & 2017 & 4.2 M & Img-class, obj-det \\\hline
MobileNet-v2 \cite{Mobilenetv2} & 2017 & 3.4 M & Img-class, obj-det \\\hline
WideResNet-40-4  \cite{WideResNet} & 2016& 8.9 M &  Img-class, obj-det \\\hline
WideResNet-16-8 \cite{WideResNet} & 2016&  11 M &  Img-class, obj-det \\\hline
WideResNet-28-10 \cite{WideResNet} & 2016& 36.5 M &  Img-class, obj-det  \\\hline

SWideRNet ($w_1$=0.25, $w_2$=0.25, $l$=0.75) \cite{SWideRNet} & 2020&   7.77 M & Panoptic-seg \\\hline
SWideRNet ($w_1$=1, $w_2$=1, $l$=1) \cite{SWideRNet} & 2020&168.77 M &  Panoptic-seg  \\\hline
SWideRNet ($w_1$=1, $w_2$=1, $l$=6) \cite{SWideRNet} & 2020&836.59 M &  Panoptic-seg   \\\hline
SWideRNet ($w_1$=1, $w_2$=1.5, $l$=3) \cite{SWideRNet} & 2020&  946.69 M &  Panoptic-seg  \\\hline

HRNet W32, W48 \cite{HRNet}&  2019 &  28.5M, 63.6M &Human-Pose- est \\\hline
HRNet V2 \cite{HRNet}&  2020  & - & Semantic-seg  \\\hline

\end{tabular}
\end{center}
\end{table*}

\subsection{Inception-ResNet-V2}
Combining ResNet and inception architecture, Szegedy et al. developed Inception-ResNet-V2 \cite{Inception-ResNet} \footnote{https://github.com/zhulf0804/Inceptionv4\_and\_Inception-ResNetv2.PyTorch} in 2016. Using residual connections (skip-connections between blocks of layers) Inception-ResNet-V2 is composed of 164 layers of 4 max-pooling and 160 convolutional layers, and about 55 million parameters. The Inception-ResNet network have led to better accuracy performance at shorter epochs. Inception-ResNet-V2 is used by many other architectures such as Faster R-CNN G-RMI \cite{23}, and Faster R-CNN with TDM \cite{24}.

\subsection{ResNeXt}

ResNet architecture contains blocks of consecutive convolutional layers while each block is connected with the previous block output. The authors in \cite{ResNeXt} \footnote{https://github.com/facebookresearch/ResNeXt} developed a new architecture based on ResNet architecture by replacing consecutive layers in each block with a set of branches of parallel layers. This model can be presented as the complex version of ResNet, which increases the model size with more parameters. But, in terms of learning, ResNeXt allows the model to learn from a set of features concatenated at the end of each block and using a variety of transformations with the same block. ResNeXt model has trained on image classification ImageNet dataset, as well as on object detection MS COCO dataset. The results are compared with ResNet results, and the obtained ResNeXt results outperform ResNet on COCO and ImageNet datasets.

\subsection{DarkNet}
In order to develop a efficient network with small size, the developer in \cite{DarkNet} introduced Darknet-19  \footnote{https://github.com/visionNoob/pytorch-darknet19} architecture based on some existing notions like inception and batch normalization \cite{17} used in GoogleNet and ResNet as well as on the notions of network In network \cite{DarkNet}. Darknet network is composed of a set of convolutional-max-pooling layers while the DarkNet-19 contains 19 convolutional. In order to reduce the number of parameters, a set of  $1\times1$ convolutional kernels is used, while  $3\times3$ convolutional kernels are not used much like in VGG of ResNet. DarkNet-19 is used for many object detection methods including  YOLO-V2, YOLO-v3-v4 \cite{27}.

\subsection{ShuffleNet}
In order to reduce the computation cost as well as conserving the accuracy, ShuffleNet  \cite{shuffleNet}\footnote{https://github.com/kuangliu/pytorch-cifar/blob/master/models/shufflenet.py} proposed in \cite{shuffleNetv2} is an computation-efficient CNN
architecture introduced for mobile devices that have limited computational power. ShuffleNet consist of point-wise group convolution and channel shuffle notions. The group convolution of depth-wise separable convolutions has been introduced in AlexNet for distributing the model over two GPUs, also used in ResNeXt an demonstrate the it robustness. For that, point-wise group convolution is exploited by ShuffleNet to reduce computation complexity of $1\times1$ convolutions. Group of convolution layers can affect the accuracy od s network, for that, the authors of ShuffleNet used channel shuffle operation to share the information across feature channels. Beside image classification and objects detection, ShuffleNet has been used as backbone for many other tasks. Also, ShuffleNet has two versions, while the second version is proposed in \cite{shuffleNetv2}.

\subsection{DetNet}
In literature we can find many object detection architecture that based on pre-trained model to detect objects. This includes SSD, YOLO family, Faster R-CNN, etc. Some of works are specifically designed for the object detection. For the luck of backbones introduced for object detection, the authors in \cite{Detnet} \footnote{https://github.com/yumoxu/detnet} proposed a deep-learning-based network. The same architecture is used also for image classification on ImageNet dataset and compared with the other famous network. For object detection, objects localization as well as the recognition of it make the process different from image classification process. For that, DetNet is introduced as a object detection backbone that consist of maintains high spatial resolution using a dilation added to each part of the network, like illustrated in Figure \ref{DetNet}.

\begin{figure}[t!]
  \centering
  \footnotesize
  \begin{tabular}[b]{c}
    \includegraphics[width=1\linewidth]{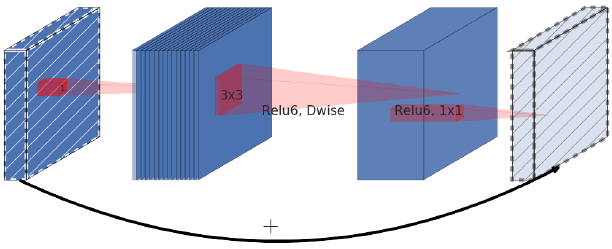} \\
    \textbf{(a) MobileNet-V2}
  \end{tabular} 

  \begin{tabular}[b]{c}
    \includegraphics[width=1\linewidth]{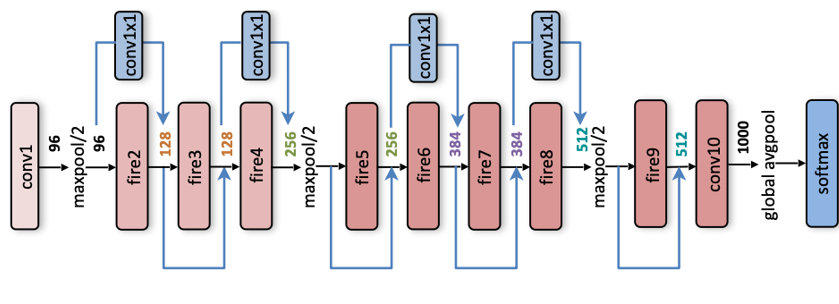} \\
    \textbf{(b) SqueezeNet}
  \end{tabular} 
  
    \caption{ MobileNet-V2 and SqueezeNet architectures.}
  \label{mobile}
\end{figure}

\subsection{SqueezeNet}
To reduce the number of parameters, the authors in \cite{SqueezeNet}\footnote{https://github.com/pytorch/vision/blob/master/torchvision/models/squeezenet.py} developed a convolutional neural network named SqueezeNet consist of using $1\times1$ filter in almost layers instead of $3\times3$ filter, since the number of parameters with 1x1 filters is 9x fewer.  Also, the input channels is decreased to $3\times3$ filters and delayed down-sampling large activation maps lead to higher classification accuracy. The SqueezeNet layers are a set of consecutive fire module. A Fire module is comprised of: a squeeze convolution layer (which has only $1\times1$ filters), feeding into an expand layer that has a mix of $1\times1$ and $3\times3$ . Like ResNet, SqueezeNet architecture has the same connection method between layers or fire module. SqueezeNet contains 1.24M parameters which is the small backbone comparing with the other networks.


\subsection{MobileNet}
In order to implement a deep learning model suitable to use it for low machine performance like mobile devices, the authors in \cite{Mobilenet} developed a model named MobileNet \footnote{https://github.com/tensorflow/tfjs-models/tree/master/mobilenet}. Depth-wise separable convolutions are used to implement MobileNet architecture which can be represented as light weight model. Here two global hyper-parameters are introduced to make the developers to choose the suitable sized model where they using it in their problem. MobileNet are trained and tested on ImageNet for image classification. 
 
Another version of MobileNet which is MobileNet-v2 proposed in \cite{Mobilenetv2} for object detection purposes. Here, the authors introduced invented residual block which allow a shortcut connection directly between the bottleneck layers like illustrated in Figure \ref{mobile}. Also, depth-wise separable convolutions are used in this version to filter features as a source of non-linearity. the architecture is trained and tested for object detection and image classification.

\begin{figure}[t!]
  \centering
  \footnotesize
 \begin{tabular}[b]{c}
    \includegraphics[width=.41\linewidth]{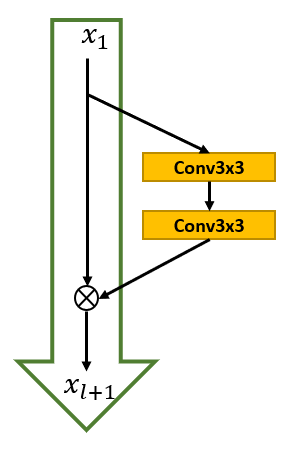}\\
    \textbf{(a) WideResNet}
  \end{tabular}
  
  \begin{tabular}[b]{c}
    \includegraphics[width=.81\linewidth]{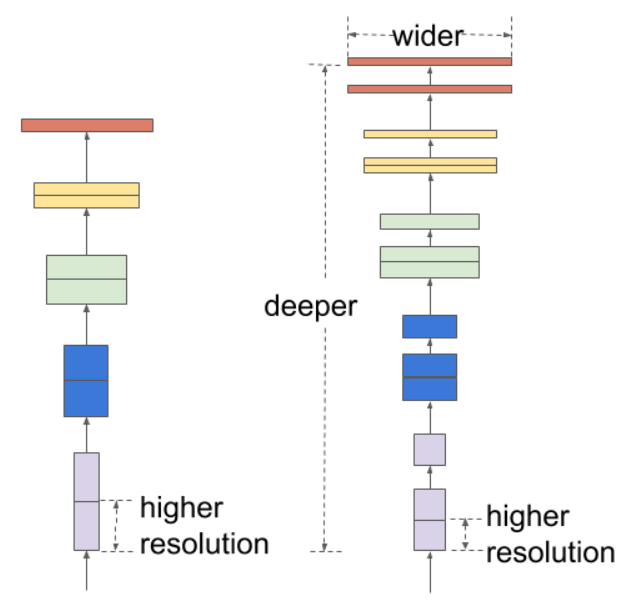}\\
     \textbf{(b) EfficientNet}
  \end{tabular}
  \caption{ WideResNet and EfficientNet architectures}
  \label{figure:wideresnet}
\end{figure}

\subsection{WideResNet}
Improving the accuracy of residual network means increasing the number of layers which make the training very slow. The researcher in \cite{WideResNet} \footnote{https://github.com/xternalz/WideResNet-pytorch} attempted to solve this limitation by proposing a new architecture named WideResNet based on ResNet blocks, but instead of increasing the depth of the network they increase the width of the network and decrease the depth. This technique allows to reduce the number of layer as well as minimizing the number of parameters. The number of parameters in WideResNet depends to number of residual layers (total number of convolutional layers) and to the widening factor $k$. WideResNet is trained and tested on CIFAR dataset unlike the other backbone which are trained and tested on ImageNet. WideResNet have demonstrated outstanding performance on image classification, object detection, and semantic segmentation.

\subsection{EfficientNet}

EfficientNet \cite{efficientnet} \footnote{https:
//github.com/tensorflow/tpu/tree/
master/models/official/efficientnet} Networks which are a recent family of architectures have been shown to significantly outperform other networks in classification tasks while having fewer parameters and FLOPs. It employs compound scaling to uniformly scale the width, depth, and resolution of the network efficiently. EfficientNet parameters being 8.4x smaller and 6.1x faster on inference than the best existing networks. Many versions of EfficientNet starting from B0 to B7. This can be easily replaced with any of the EfficientNet models based on the capacity of the resources that are available and the computational cost. EfficientNet-B0 contains 5.3 million parameters while the last version  EfficientNet-B7 has 66M parameters .

\subsection{SWideRNet}
Scaling Wide Residual Network (SWideResNet) is a new network developed for image segmentation \cite{SWideRNet}, obtained by incorporating the simple and effective Squeeze-and-Excitation (SE) and Switchable Atrous Convolution. Using the same principal of WideResNet, SWideResNet also adjusting the number of layers as well as the channel size (depth) of the network. In addition, SWideResNet used SAC block instead of simple convolutional layers like WideResNet. Also, the Squeeze-and-Excitation block is added after each stage of the network. The number of parameters of SWideResNet is depend to the scales of channels of the first two stages($w_1$)  of the network and of the remaining stages($w_2,l$) like presented in Table 1. 

\subsection{Xception}

Inspired by Inception network, Xception \cite{Xception} is another backbone used for features extraction. Xception is a depth-wise separable convolution added to Inception module with a maximally large number of towers. Where Inception V3 modules have been replaced with depth-wise separable convolutions. The proposed network is used in many computer science and vision tasks including object detection crowd counting, and images segmentation, etc.

\begin{figure*}[t!]
  \centering
  \footnotesize
     \begin{tabular}[b]{c}
    \includegraphics[width=.8\linewidth]{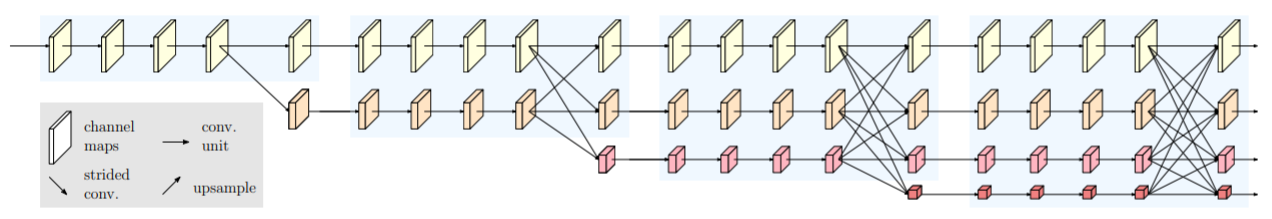}\\
  \end{tabular}
  \caption{ HRNetV2 architecture}
  \label{figure:HRNetV2}
\end{figure*}

\begin{figure}[t!]
  \centering
  \footnotesize
  \begin{tabular}[b]{c}
    \includegraphics[width=.5\linewidth]{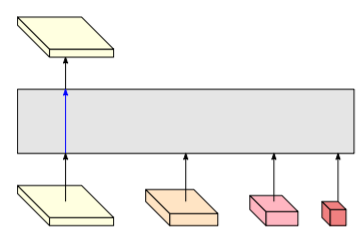} \\
    \textbf{HRNetV1 }
  \end{tabular}

    \begin{tabular}[b]{c}
    \includegraphics[width=.5\linewidth]{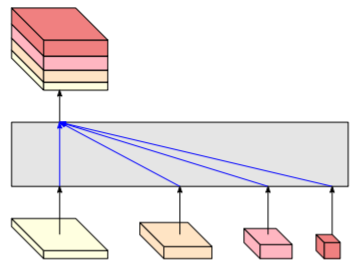} \\
    \textbf{HRNetV2}
  \end{tabular} 
 
  \caption{ HRNetV1 and HRNetV2 representation of the fusion strategy}
  \label{HRNet}
\end{figure}

\subsection{HRNet}  \cite{HRNetv2}
To maintain the high resolution of an images during the training process, the authors in \cite{HRNet} proposed a High-Resolution Net (HRNet) that consist of starting by high-resolution sub-network then move to to-low resolution sub-network and so on, using multiple stage with the same scenario. The multi-resolution sub-networks are connected parallely before fusion them. HRNet is used at the first time for human pose estimation then the second version of HRNet-v2 is proposed and used for semantic segmentation.. For the HRNetV2 the difference is in fusion operation between multi-resolution sub-networks like presented in Figure \ref{HRNet} like illustrated in Figure \ref{figure:HRNetV2}. While for HRNet-v1 the representation from the high-resolution convolution stream is take into account. For HRNetV2: Concatenate the (up-sampled) representations that are from all the resolutions.

\section{Tasks related to Backbones}

\subsection{Image classification }

Image classification was and still one of the hot topics in computer vision using deep-learning-based techniques. It was the subject of evaluation of many deep-learning-based models. Due to the evolution of deep convolutional neural networks (CNNs), the performance of image classification methods has become more accurate and also faster \cite{27}. The evaluation of CNN-based image classification methods used the famous dataset in this task which is ImageNet. These networks have also been references for other topics due to the novelty of each one of them, as well as the evaluation of these networks on many datasets \cite{28}. From these backbones or reference networks, we can find all the cited networks including VGGs, ResNets, DenseNet, and the others.

The efficiency of such network generally related to the accuracy of it. Here the image classification evaluation has been performed on ImageNet for all these proposed networks. But, we can find also another characteristic which can be considered for comparing the models, which is the computational complexity. For image classification, Floating Point Operations Per Second (FLOPs) is the performance measurement metrics for computing the complexity in term of speed and latency of a model \cite{29}. The existing models that have the best accuracy, in general, have a high FLOPs values. For example, the researcher impalement some models used by the devices with limited computing power. From these models we can find MobileNet that implemented for smartphones as well as ShuffleNet and Xception. Another factor that can affect the speed of a model is the parallelism. For example, under the same FLOPs a model can be faster that another while the degree of parallelism is higher.

Image classification task can be considered as simplest computer vision task in terms of information extracted from the images \cite{30}. This is compared with the other computer vision tasks like panoptic segmentation, object tracking, Action recognition, etc. But, it can presented also as the test room of any deep-learning-based model. A comparison between the famous networks for images classification is presented in discussion section.

\begin{figure*}[t]
\centering
\begin{tabular}[b]{c}
    \includegraphics[width=1\linewidth]{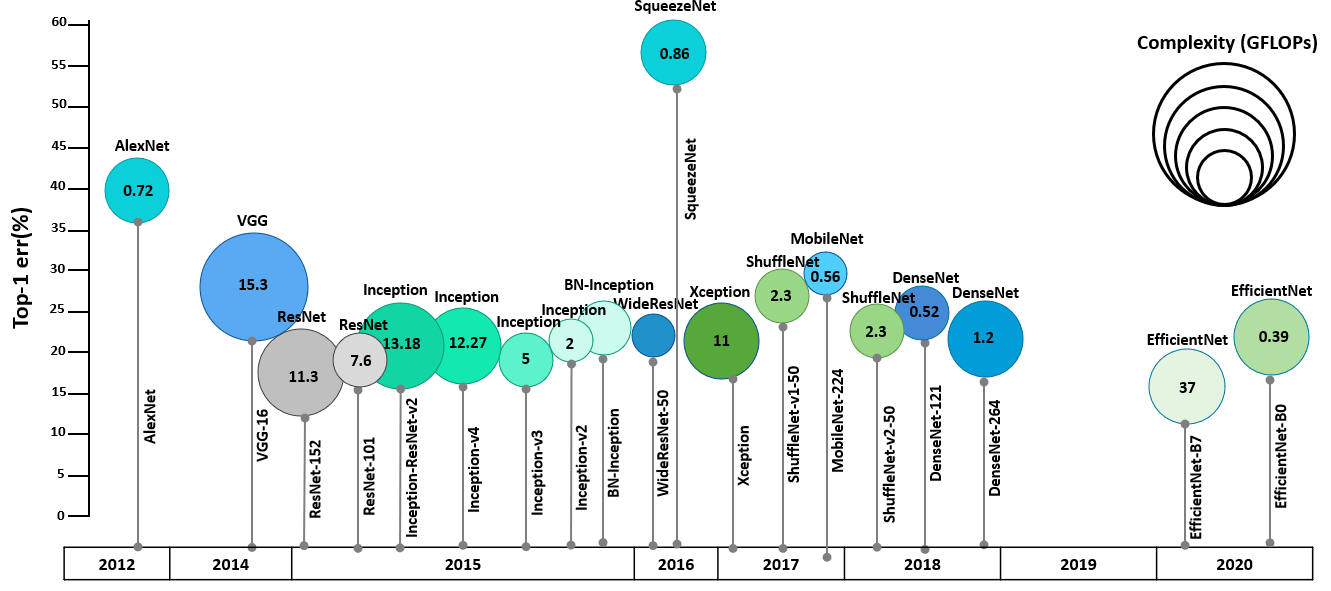} \\
\end{tabular} 
\caption{ Timeline and performance  accuracies of the the proposed networks on ImageNet.}
\label{fig:Appblck}
\end{figure*}

\subsection{Object detection}
Object detection and recognition is one of the hot topics in computer vision  \cite{o15}. Many challenges can be found for detecting the objects in an images including the scale of the objects, the similarity between some objects, and thee overlapping between them \cite{o16}. Object detection models that are optimal for detection need to have a higher input network size for smaller object. Multiple layers are needed to cover the increased size of input network for a higher receptive field. To detect multiple objects there’s a need for different sizes in a single image \cite{o22}. The following are different models used for object detection literature \cite{o17}.

\subsubsection{YOLOV3} YOLO-v3 is from the series of YOLO object detection models where YOLO-v3 is the third version. YOLO is short form for ”You Only Look Once”. YOLO detects multiple objects at a time. Its based on convolutional Neural Network, predicts classes as well as localization of the objects. Applying single Neural Network, it divides the image into grid cells and from this cell probabilities are generated. It predicts the bounding boxes using anchor boxes and outputs the best bounding box on the object. YOLO-v3 over here consists of 53 convolutional layers also called Darknet-53, for detection it has 53 more layers having a total of 106 layers. The detection happens at layers 82, 94 and 106 \cite{yolov3}. Images are often resized in YOLO according to input network size to improve detection at various resolutions. It predicts offsets to bounding boxes by normalizing the bounding box coordinates to width and height of image to eliminate gradients. The score represents the
probability of predicting the object inside the bounding box. Predicted probability to the Intersection of the union that is measure of predicted bounding box to the ground truth
bounding box \cite{o17,o22}. This model was extensively used in many forms by \cite{o17} for training and testing and as a detector in the YOLOv4 architecture. 

\subsubsection{YOLO-v4}
YOLO-v4 produces optimal speed and accuracy compared to YOLO-v3. It is a predictor with a backbone, neck, dense prediction and sparse prediction. In the backbone several architecture can be used like Resnet, VGG16 and Darknet-53 \cite{yolov4}. The neck enhances feature discriminability by methods like Feature pyramid network (FPN), PAN and RFB. The head handles the dense prediction using either RPN, YOLO or SSD. For the backbone, Darknet-53 was tested to be the most superior model \cite{yolov4}. YOLO-v3 is commonly used as head for YOLO-v4. To optimise training, data augmentation happens in the backbone. Its proven to be faster and more accurate than YOLO-v3 with MS COCO dataset . Advantage of YOLO-v4 is that it runs on a single GPU ,enabling less computation. With its wide variety of features. It achieved 43.5\% for mAP metric on COCO dataset with approximately 65 FPS.

\subsubsection{YOLO-v4-tiny}
YOLO-v4-tiny is a compressed version of YOLO-v4 with its network size decreased having lesser number of convolutional layers with CSPDarknet backbone. The layers are reduced to three and anchor boxes for prediction is also reduced enabling faster detection. 

\subsubsection{Detectron}
An object detection model built with Pytorch, has efficient training capability. Its Mask R-CNN, FPN–R50 benchmarks has a modular architecture, the input goes through a CNN backbone to extract features, which are used to predict region proposals. Regional features and image features are used to predict bounding boxes \cite{Detectron}. The scalability of this model and region proposal feature enables accurate detection.

\subsubsection{YOLO-v5} 
YOLO-v5 is a pytorch implementation of an improved version of YOLO-v3, published in May 2020 by Glenn Jocher of Ultralytics LLC5 on GitHub6 ~\cite{o30}. It is an improved version of their well known YOLO-v3 implementation for PyTorch. It has similar implementation to YOLO-v4, where it incorporates several techniques like data augmentation and changes to activation function with post processing to the YOLO architecture. It combines images for training and uses a self-adversarial training (SAT) claiming an accelerated inference ~\cite{o30,o31}.

\subsection{Crowd counting}

Crowd counting is the operation of estimating the number of people or objects in a surveillance scene. For people counting in the crowd, many works have been proposed for estimating the crowd mass. The proposed methods can be divided into many categories such as regression-based methods, density estimation-based methods, detection-based methods, and deep-learning-based methods. Comparing the accuracy of each one of these categories the CNN-based methods are the most effective methods. 

The introduction of deep learning techniques makes the computer vision tasks more effective and the Convolutional Neural Network (CNN) improves the performance accuracy of each task specialty those used large-scale datasets. On crowd counting, the use of deep learning techniques allows the estimation of crowd density more accurate comparing with the traditional and sequential methods in terms of accuracy and the computational cost \cite{count1,count2} . Also, the used backbones and the interconnection between the part of a network has an impact on the accuracy of a CNN model. Different backbones including VGG-16, VGG-19, ResNet-101, and others have been used in different crowd counting models, but the most used backbone for crowd counting is VGG-16. The use of these backbones can increase the consumption cost, especially on large-scale datasets. Also using VGG-16 for feature extraction, the authors in \cite{count4} proposed a crowd counting method named DENet composed of two-stage networks: detection network (DNet) and estimation network (ENet). Detection network count the people in each region and the estimation network ENet work on the complex and crowded regions in the images. Using the same backbone, another method has been proposed named CANNet for estimating the crowd density map \cite{count5}.

Also in \cite{count6} the authors based on the contextual and spatial information of the image as well as VGG-16 for crowd counting method implementation. The method named SCAR consists of a Spatial-wise attention module and a Channel-wise Attention module before combining the results of each module for the final estimation.
In the same context, the authors in \cite{count21} proposed an adaptive dilated self-correction model (ADNet) which estimates the density. Using a multitask model that consists of two proposed networks - Density Attention Network (DANet) and Attention Scaling Network (ASNet) \cite{count22}. The authors estimate the attention map which is a segmentation of the crowd regions before using this result to estimate the density of the crowd. For the two methods \cite{count21} and \cite{count21} VGG-16 is also used as backbone.

Using another version of VGG family which is VGG-19, the authors in \cite{count7} proposed a method based on density probabilities construction and Bayesian loss function (BL) to estimate the crowd density maps. In the same context, the authors in \cite{count8} proposed a crowd counting dataset as well as a crowd counting method based on a Special FCN model (SFCN). This method used ResNet-101 as backbone, which is the from a few method used ResNet.  

In order to reduce the number of parameters and the size of a network,  the authors used MobileNet-v2 backbone to reduce the FLOPs and modeling a Light-weight encoder-decoder crowd counting model \cite{count3}.

\subsection{Video summarization}
 Currently, real-time useful information extraction from the video is a challenge for different computer vision applications, especially from large videos. The extracted information allows reducing the time of searching as well as allowing to detect and identify some features that can useful to be used in other tasks. The summarization of videos is one of the most techniques applied to extract this information. Summarizing useful information from videos has been the main purpose of many studies recently \cite{vs0}. It's considered as an important step to improve the video surveillance systems in terms of reducing the searching time for a specific event as well as simplifying the analysis of a huge number of data. To do this, many aspects can be considered for a good summarization such as the type of scene analyzed (Private or public, indoor or outdoor, crowded or not). Also, the pre-processing can be used for enhancing the summarization process which supposed to be with less space for storage in less computational time \cite{vs01}.

Using deep learning and deep reinforcement learning techniques many methods have been proposed. These methods used many backbones in their models for features extraction. For example, in \cite{vs1} the authors proposed a language-guided video summarization using a conditional CNN-based model while GoogleNet and ResNet backbones are used for features extraction. The same backbone GoogleNet and ResNet-50 are used in \cite{vs2} for multi-stage networks for video summarization. In the same context, and using Sparse Autoencoders network with Random Forest Classifier, the authors in \cite{vs3} proposed a CNN-based model for key-frames selection. A set of backbones, including AlexNet, GoogleNet, VGG-16, and Inception-ResNet-v2, have been used then compared the impact of each one on the summarization results using VSUMM and OVP datasets. In \cite{vs4}, the ConvNet network is used for features extraction of the proposed video summarization model. By scene classification for video summarization, the authors in \cite{vs5} used many backbones for features extraction including  VGG-16, VGG-19, Inception-v3, and ResNet-50. To summarize the daily human behaviors, the authors in \cite{vs55} proposed a deep learning method using ResNet-152 backbones. Using the same backbone, object-based video summarization with a DRL model has been proposed in \cite{vs10} to summarize the video based on the detected object in it. The objects are tracked using an encoder-decoder network before selecting the target object to be summarized. 

Using deep reinforcement learning, many methods have been proposed for video summarization using different backbones \cite{vs6,vs7,vs8,vs9,vs11,vs12,vs13}. Highlight or summarize a video can be different from a user (person) to another. Existing deep learning methods attempted to summarize the videos using different models but with one point of view. In other to overcome this limitation, as well as to detects different highlights according to different user’s preferences, deep reinforcement learning has been used in \cite{vs6}. The algorithm used a reward function to detect each user performance as a highlight candidate, also used ResNet-50 as backbone. 

An effective video summarization method should analyze the semantic information in the videos. By using DRL for selecting the most distinguishable frames in a video, the authors in \cite{vs7} cut the videos using the Action parsing technique. Here, AlexNet has been used for features extraction. Using a weakly supervised hierarchical reinforcement learning architecture exploited GoogleNet for features extraction, the authors in \cite{vs8} select the representative video key-frames as summarization results which represents also a video shot after collecting them together. 
In addition to the content-based video summarization, some researchers proposed a query-conditioned video summarization to summarize the video based on a given query \cite{vs9}.  The proposed method named Mapping Network (MapNet) used two users query as inputs and the DRL-based framework provide two different summaries for the requested queries. For features extraction, ResNet-152 is used. For summarizing a video into key-frames that represents
the contextual meaning of the video, the authors in \cite{vs11},
\cite{vs12} proposed Deep Summarization Network(DSN) method
based on a new deep reinforcement learning reward function that accounts the representativeness and diversity of each frame of the video. Two methods used GoogleNet as a backbone. In the same context, using the semantic of the videos the authors in \cite{vs13} proposed Summary Generation Sub-Network (SGSN) to select the
key-frames from a given video using a DRL reward function and Inception-v3 as a backbone.


\begin{table*}
\begin{center}
\scriptsize
\caption{Various methods and Backbone used for each task. }
\label{backbone-table}
\begin{tabular}{|c|l|c|c|l|c|}
\hline
\multicolumn{1}{c}{\textbf{Task}}  &\textbf{Method} &	\textbf{Backbone}&\multicolumn{1}{|c|}{\textbf{Task}}  &\textbf{Method} &	\textbf{Backbone} \\ \hline

\multirow{15}{*}{\rotatebox[origin=c]{90}{\textbf{Image classification}} }& AlexNet \cite{AlexNet} &AlexNet &
\multirow{13}{*}{\rotatebox[origin=c]{90}{\textbf{Object detection}} }& MobileNet \cite{Mobilenet} & MobileNet  \\\cline{2-3}\cline{5-6}

&  GoogLeNet & GoogLeNet   &&   ShuffleNet \cite{shuffleNet}  & ShuffleNet	\\\cline{2-3}\cline{5-6}
& VGG  \cite{VGG} & VGG  &&   Xception  \cite{Xception} & Xception \\\cline{2-3}\cline{5-6}
&  ResNet \cite{ResNet} &ResNet    &&   MobileNet-v2 \cite{Mobilenet} & MobileNet-v2 	\\\cline{2-3}\cline{5-6}
&  DenseNet\cite{DenseNet} & DenseNet   &&   DetNet \cite{Detnet} & DetNet	\\\cline{2-3}\cline{5-6}
& DetNet \cite{Detnet} &DetNet   && SSD \cite{ssd} & VGG-16	\\\cline{2-3}\cline{5-6}
&  SqueezeNet \cite{SqueezeNet}& SqueezeNet   &&   ShuffleNet-v2 \cite{shuffleNetv2} & ShuffleNet-v2	\\\cline{2-3}\cline{5-6}
&  ResNeXt \cite{ResNeXt}& ResNeXt    &&   WideResNet \cite{WideResNet}& WideResNet	\\\cline{2-3}\cline{5-6}
&  Xception \cite{Xception}& Xception   &&  RetinaNet \cite{retinanet} &ResNet-101	\\\cline{2-3}\cline{5-6}
&  BN-Inception \cite{BNInception}& BN-Inception    &&  RetinaNet \cite{retinanet} &ResNeXt-101	\\\cline{2-3}\cline{5-6}
&  Inception-v2 \cite{Inceptionv3}&Inception-v2   &&   Faster R-CNN G-RMI \cite{23}&Inception-ResNet-v2 	\\\cline{2-3}\cline{5-6}
&  Inception-v3 \cite{Inceptionv3}& Inception-v3  &&   Faster R-CNN TDM\cite{24}& Inception-ResNet-v2	\\\cline{2-3}\cline{5-6}
&Inception-ResNet-v1 \cite{Inception-ResNet} & Inception-ResNet   &&   YOLOV2 \cite{27} & DarkNet-19	\\\cline{2-3}\cline{5-6}
&  Inception-v4 \cite{Inception-ResNet} & Inception-v4   &&   YOLO-V3 \cite{yolov3} & DarkNet-19	\\\cline{2-3}\cline{5-6}
&  Inception-ResNet-v2 \cite{Inception-ResNet} & Inception-ResNet-v2   &&  YOLO-V4 \cite{yolov4}& CSPDarknet53	\\\cline{2-3}\cline{5-6}
&  WideResNet \cite{WideResNet} & WideResNet   &&   EfficientDet \cite{EfficientDet}& EfficientNet	\\\cline{2-3}\cline{5-6}
& MobileNet \cite{MobileNet}& MobileNet   &&  DetectoRS \cite{Detectors}&ResNet-50 	\\\cline{2-3}\cline{5-6}
&  ShuffleNet-v2 \cite{shuffleNetv2} & ShuffleNet-v2   &&   DetectoRS \cite{Detectors}& ResNeXt-101	\\\cline{1-3}\cline{4-6}\hline

\multirow{15}{*}{\rotatebox[origin=c]{90}{\textbf{Panoptic segmentation}} }& FPSNet \cite{PS67} &	ResNet-50 &
\multirow{13}{*}{\rotatebox[origin=c]{90}{\textbf{Crowd counting}} }& CSRNet \cite{count1} (2018) & VGG-16  \\\cline{2-3}\cline{5-6}
&Axial-DeepLab \cite{PS68} &	DeepLab  &&SPN  \cite{count2} (2019)	&VGG-16	\\\cline{2-3}\cline{5-6}
&BANet \cite{PS69} &	ResNet-50 &&DENet\cite{count4} 	&VGG-16	\\\cline{2-3}\cline{5-6}
&VSPNet \cite{PS94} &	ResNet-50  &&CANNet \cite{count5}	&VGG-16	\\\cline{2-3}\cline{5-6}
&BGRNet \cite{PS95} &	ResNet50  &&SCAR \cite{count6}   &VGG-16\\\cline{2-3}\cline{5-6}
&SpatialFlow \cite{PS96} &	ResNet50  &&ADNet  \cite{count21}  & VGG-16  \\\cline{2-3}\cline{5-6}
&Weber et al. \cite{PS97} &	ResNet50  &&ADSCNet  \cite{count21}  &  VGG-16  \\\cline{2-3}\cline{5-6}
&AUNet \cite{PS99} &	ResNet50  &&ASNet  \cite{count22} & VGG-16  \\\cline{2-3}\cline{5-6}
&OANet \cite{PS100} &	ResNet50  &&SCNet  \cite{count9} & VGG-16 \\\cline{2-3}\cline{5-6}
&SPINet \cite{PS102} &	ResNet50  &&BL  \cite{count7} &VGG-19 \\\cline{2-3}\cline{5-6}
&Son et al. \cite{PS107} &	ResNet-50  &&MobileCount \cite{count3} & MobileNet-V2 \\\cline{2-3}\cline{5-6}
&SOGNet \cite{PS98} &	ResNet101  &&SFCN  \cite{count8} & ResNet-101\\\cline{2-3}\cline{4-6}\cline{4-6}

&DR1Mask \cite{PS103} &	ResNet101  &\multirow{18}{*}{\rotatebox[origin=c]{90}{\textbf{Video summarization}}}& GoogleNet+Transformer \cite{vs1}&  GoogleNet\\\cline{2-3}\cline{5-6}

&DetectoRS \cite{Detectors} &ResNeXt-101 &&ResNet+Transformer \cite{vs1}&  ResNet\\\cline{2-3}\cline{5-6}

&EfficientPS \cite{PS91}&	EfficientNet  &&\multirow{2}{*}{MCSF \cite{vs2}}&GoogleNet\\\cline{2-3}\cline{6-6}

&EffPS-b1bs4-RVC \cite{PS25} &	EfficientNet-B5  &&&ResNet\\\cline{1-3}\cline{5-6}\cline{1-3}

\multirow{15}{*}{\rotatebox[origin=c]{90}{\textbf{Action recognition}}}&Yang et al. \cite{ar01}&  ResNet-50 & &\multirow{3}{*}{Nair et al. \cite{vs3}}&AlexNet \\\cline{2-3}\cline{6-6}

&TEA  \cite{ar02}& ResNet-50  & &&GoogleNet\\\cline{2-3}\cline{6-6}

&\multirow{2}{*}{Sudhakaran et al.  \cite{ar03}} &  BN-Inception & &&VGG-16  \\\cline{3-3}\cline{6-6}
&& Inception-v3 & &&Inception-ResNet-v2\\\cline{2-3}\cline{5-6}
&MM-SADA  \cite{ar04}&3D ConvNet & &\multirow{3}{*}{Rafiq et al. \cite{vs4}}&VGG-16\\\cline{2-3}\cline{6-6}

&I3D \cite{ar05}&  3D ConvNet & && VGG-19\\\cline{2-3}\cline{6-6}

&\multirow{3}{*}{Li et al.  \cite{ar6}}& ResNeXt-101 & &&Inception-v3\\\cline{3-3}\cline{6-6}
&& ResNet-18 & &&ResNet-50\\\cline{3-3}\cline{5-6}

&& ResNet-152 & &Zhang et al. \cite{vs5} & ResNet-152 \\\cline{2-3}\cline{5-6}
&Wu et al. \cite{ar1}& ConvNet&& Wang et al. \cite{vs6}& ResNet50\\\cline{2-3}\cline{5-6} 
&PEAR  \cite{ar2}&  BN-Inception  & &Lei et al. \cite{vs7}& AlextNet\\\cline{2-3}\cline{5-6}

&Chen et al.  \cite{ar3} &  VGG-16 & &Chen et al. \cite{vs8}& GoogleNet\\\cline{2-3}\cline{5-6}

&Li et al.  \cite{ar4}& GoogleNet & &Zhang et al. \cite{vs9}& ResNet-152\\\cline{2-3}\cline{5-6}

&\multirow{2}{*}{Dong et al. \cite{ar5}}& BN-Inception & & Zhang et al. \cite{vs10} & ResNet-152\\\cline{3-3}\cline{5-6}

&& ConvNet & &DR-DSN \cite{vs11}& GoogleNet   \\\cline{2-3}\cline{5-6}

&Wang et al.  \cite{ar6}&GoogleNet & &DR-DSN-s \cite{vs11}& GoogleNet  \\\cline{1-3}\cline{5-6}


\multirow{11}{*}{\rotatebox[origin=c]{90}{\textbf{COVID-19 detection}}}&  \cite{cov7}, \cite{cov14}, \cite{cov38} & VGG16& & Wang et al. \cite{vs12}&GoogleNet\\\cline{2-3}\cline{5-6}

&\cite{cov5}, \cite{cov7}, \cite{cov11}, \cite{cov15}, \cite{cov24} &  VGG19& &SGSN \cite{vs13}  & Inception-V3   \\\cline{2-3}\cline{4-6}
&\cite{cov13}, \cite{cov14}, \cite{cov24}, \cite{cov28}, \cite{cov35}&   ResNet-18& \multirow{10}{*}{\rotatebox[origin=c]{90}{\textbf{Face recognition}}}&CosFace(LMCL) \cite{fra12}	& ConvNet \\\cline{2-3}\cline{5-6}

&\cite{cov13}, \cite{cov29}, \cite{cov33}, \cite{cov34}, \cite{cov39}&   ResNet-50& &Range loss \cite{fra13} 	& VGG-19
 \\\cline{2-3}\cline{5-6}

&\cite{cov7}, \cite{cov14}, \cite{cov15}, \cite{cov33}, \cite{cov34} &   Inception-v3& & NRA+CD \cite{fra15} &ResNet-50 \\\cline{2-3}\cline{5-6}

&\cite{cov15}, \cite{cov34}   & Inception-ResNet-v2& &RegularFace \cite{fra17} 	& ResNet-20 \\\cline{2-3}\cline{5-6} 

&\cite{cov7},\cite{cov11}, \cite{cov14}, \cite{cov15}, \cite{cov29} &   Xception& &Range loss \cite{fra13} 	& VGG-19\\\cline{2-3}\cline{5-6} 

&\cite{cov17}, \cite{cov28}& AlexNet  & &ArcFace \cite{fra18} &ResNet-100 \\\cline{2-3}\cline{5-6} 

&\cite{cov28},\cite{cov33}, \cite{cov35}   &  GoogleNet   & &FairLoss  \cite{fra19} &ResNet-50\\\cline{2-3}\cline{5-6} 

&\cite{cov13}, \cite{cov15}, \cite{cov24}, \cite{cov33}&   DenseNet   & &Attention-aware-DRL \cite{fra20} & ResNet\\\cline{2-3}\cline{5-6}

&\cite{cov11},\cite{cov15}, \cite{cov24}, \cite{cov38} &   MobileNet-v2   & &Margin-aware-DRL \cite{fra21} & ResNet-50\\\cline{2-3}\cline{5-6}

&\cite{cov13}, \cite{cov24}, \cite{cov30}&  SqueezeNet  & &Skewness-aware-DRL \cite{fra22}&ResNet-34\\\cline{2-3}\cline{5-6}

\hline
\end{tabular}
\end{center}
\end{table*}

\subsection{Action recognition} 

Action recognition aims to recognize various actions from a video sequence under different scenarios and distinct environmental conditions. This is a very challenging topic in computer vision due to (i) its wide range of applications, including video surveillance, tracking, health care, and human-computer interaction, and (ii) its corresponding issues that require powerful learning methods to achieve a high recognition accuracy. To that end, deep learning (DL) and deep reinforcement learning (DRL) has been investigated in several frameworks for different purposes, e.g. action predictability to perform early action recognition (EAR), video captioning, and trajectory forecasting. 

For deep-learning-based methods, many architectures have been proposed using different backbones for feature extraction. For example, the authors in \cite{ar01} proposed a temporal pyramid network for recognizing human action. The proposed method consists of classifying the same action that has variant tempos. This method is implemented using 3D ResNet. The same backbone has been used in \cite{ar02} for an action recognition named Temporal Excitation and Aggregation (TEA). For the same purpose,  BN-inception and Inception-v3 backbone have been used in \cite{ar03}, while the authors proposed an action recognition method for learning the powerful representations in the joint Spatio-temporal feature space in a video. Using a multi-modal action recognition method on skeleton data, the authors in \cite{ar04} used 3D ConvNet as a backbone. The same backbone has been investigated in \cite{ar05} for the same purpose. For action detection and recognition, the authors in \cite{ar06} proposed a Spatio-temporal attention model using ResNet-152 backbone for features extraction.

Action recognition DRL-based methods are also exploited different backbones for features extraction. For example in \cite{ar1}, the authors used ConvNet for extraction features exploited for the proposed action recognition DRL-based method. Accordingly, because of the absence of fine-grained supervision, the authors in \cite{ar2} uses a DRL-based technique for optimizing the evaluator, which is fostered by recognizability rewards and early rewards. In this context, EAR is achieved using predictability computed by the evaluator, where the classifier learns discriminative representations of subsequences. In this method, BN-Inception is used as a backbone. 

In  \cite{ar3}, due to the fact that (i) using the evolution of overall video frames for modeling actions can not avoid the noise of the current action, and (ii) losing structural information of human body reduces the feature capability for describing actions; the authors introduced a part-activated DRL approach to predict actions using VGG-16 backbone. 

On the other hand, some works have focused on the use of visual attention to perform action recognition due to its benefit of reducing noise interference. This is possible by concentrating on the pertinent regions of the image and neglecting the irrelevant parts. Accordingly, in \cite{ar4}, a deep visual attention framework is proposed using DRL and GoogleNet for features extraction, in which RNN with LSTM units are deployed as a learning agent, while DRL is utilized for learning the agent’s decision policy. In the same manner, in \cite{ar5}, the irrelevant frames are ignored and only the most discriminative ones are preserved through using an attention-aware sampling scheme. Indeed, the mechanism of the key-frames extraction from videos is formulated as a Markov decision process. In this method, two backbone has been used such as BN-Inception and ConvNet. On the other side, in  \cite{ar6}, starting from the fact that attention mechanism might mimic the human drawing attention process before selecting the next location to focus(i.e. observe analyze and jump instead of describing continuous features), the authors present a framework relying on designing a recurrent neural network-based agent with GoogleNet backbone, which selects attention regions using DRL at every timestamp. 


\subsection{Face recognition}
The last decades have shown an increased interest in computer vision tasks including face recognition due to the technical development in video surveillance and monitoring.  Face recognition is able to recognize uncooperative subjects in a nonintrusive manner compared with other biometrics like fingerprint, iris, and retina recognition. Thus, it is applicable in different sectors, including border control, airports, train stations, and indoor like in companies or offices. Many works have been reported with high performance. Some of these methods have been incorporated with surveillance cameras for person identification exploiting large-scale datasets. Also, analyzing the face is the main task for several biometric and non-biometric applications including the recognition of facial expression \cite{fr9}, face identification in images under low-resolution \cite{fr10}, and face identification and verification under pose variations, which represents the most subjects focusing on the analysis of the face. 

The data structure can be considered to improve the learning, which not the case for almost all face recognition proposed methods. Deep face recognition methods use different loss functions to improve the classification results \cite{fra12,fra13,fra15,fra17,fra18,fra19,fra191}. Sofmax loss is one of the effective models for CNN-based face recognition. Other methods, combine different features Softmax to enhance the recognition, including Sofmax+contrastive, Softmax Loss+Contrastive , L-Softmax Loss , Softmax+Center Loss, and Center Loss. Where other approaches s use some loss function methods like Triplet Loss, Range loss \cite{fra13}, CosFace(LMCL) \cite{fra12}, FairLoss \cite{fra19}. Our method provides a new data structure on which the proposed method can be performed with any loss function while maintaining high accuracy.

Using DRL-based techniques face recognition was the subject of many studies. For example, to use DRL for recognizing the face, the authors in \cite{fra20} proposed an Attention-aware-DRL-based approach to verifying the face in a video. To finding the face in a video, the proposed model is formulated as a Markov decision process. The image space and the feature space are used to train the proposed model, unlike the existing deep learning models that used one of them. Another researcher paper that exploited DRL techniques is proposed in \cite{fra21}. The method introduced margin-aware RL techniques exploiting three loss functions including angular softmax loss (SphereFace), large margin cosine loss (CosFace), and additive angular margin loss (ArcFace). LFW and YTF datasets are used to train the proposed Fair-Loss method. In the same context, Wang et al. \cite{fra22} introduced Rl-race-balance-network (RL-RBN) for face recognition. Finding the optimal margins for non-Caucasians is a process formulated as a Markov decision process and exploit Q-learning to make agents learn the selection of the appropriate margin. The proposed method used ResNet34 as a backbone and RFW dataset for their learning process.

\begin{figure*}[t!]
  \centering
    \includegraphics[width=.7\linewidth]{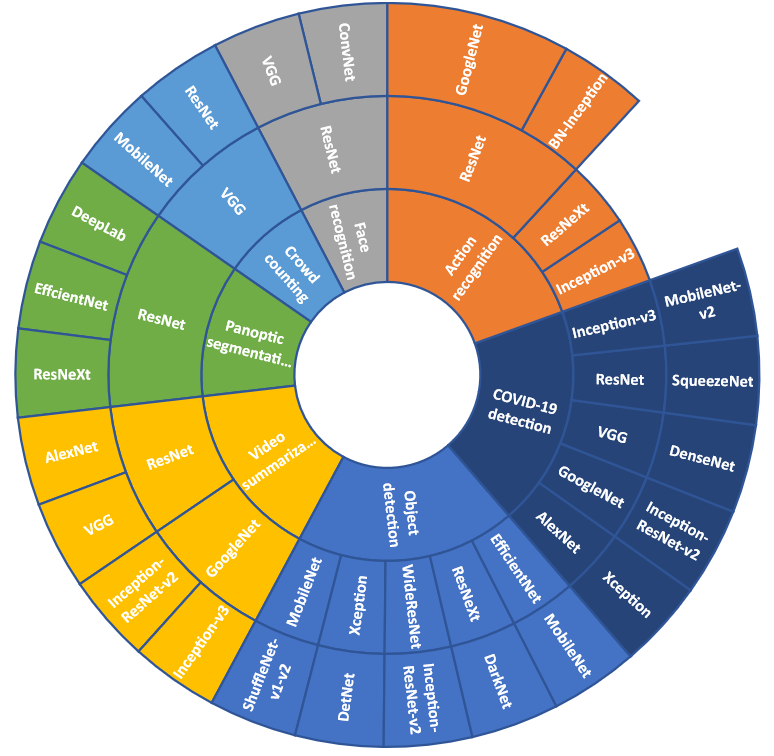} \\
  \caption{ Used backbones for each task. In some tasks some backbones are used more than the others and are illustrated with large scale in the figure.}
  \label{figure:backbone-cerle}
\end{figure*}

\subsection{COVID-19 detection}

Early work in COVID-19 detection is to extract images on patient lungs using the ultrasound technology, a technique to identify and monitor patients affected by viruses. Therefore, the development of detection and recognition techniques is needed which is  capable of automating the process without needing the help of skilled specialists \cite{cov1,cov2,cov3,cov4}. From these techniques, we can find computer vision based techniques that help in detection using images and videos. The study in \cite{cov5}, demonstrated how deep learning could be utilized to detect COVID-19 using images no matter what is the source, X-Ray, Ultrasound, or CT scan . They built a CNN model based on a comparison of several known CNN models. Their approach aimed to minimize the noises so that the deep learning uses the image features to detect the diseases. This study showed better results in ultrasound images compared to CT scans and x-ray images. They used  VGG19 network as backbone in their detection model. In the same context, Horry et al. \cite{cov7} proposed a model to detect COVID-19. The system consists of four backbone such as VGG, Xception, Inception and ResNet.

The accuracy of such method is related to the annotated data by the experts and the deep learning models used which have the potential for COVID-19 detection. For that, techniques to detect COVID-19 using the concept of transfer learning were proposed with five variants of CNNs. VGG-19, MobileNet-v2, Inception, Xception , and ResNet-v2 are used in the first experiment and MobileNet-v2 in a second assay \cite{cov11}.  Minaee et al. \cite{cov13} proposed a method named Deep-COVID based on the concept of deep transfer learning to detect COVID-19 from x-ray images. The authors used different backbones such as  ResNet-18, ResNet50, SqueezeNet, and DenseNet-121. SqueezeNet technique gives the best performance in their experiment. 

In another research paper, Moutounet et al. \cite{cov14} developed DL schema to differentiate between COVID-19 and other pneumonia from x-ray images. While VGG-16, VGG-19, Inception-ResNet-v22, InceptionV3, and Xception techniques  are invested in their diagnosis. The best performance was obtained using VGG-16. Recently, the authors in \cite{cov15} developed a CNN variants schema to detect COVID-19 from X-Ray images. The experiments proved that the VGG-19 and DenseNet were the best networks to detect Covid-19. In the same context, and to identify COVID-19, Maguolo and Nanni \cite{cov17} tested the AlexNet technique with 10 fold cross validation for training and testing. While  Chowdhury et al. \cite{cov24} used the transfer learning with the image augmentation techniques to train and validate some pre-trained networks. Furthermore , it was proposed \cite{cov25} a modified CNN to detect coronavirus from x-ray images. They combined Xception, ResNet50-V2 techniques and applied 5-fold cross-validation techniques to classify three class including COVID-19.

Loey et al. \cite{cov28} presented a deep transfer learning model with  Generative Adversarial Network (GAN) to detect COVID-19. Three backbones were tested such as AlexNet, GoogleNet, and ResNet-18. Furthermore , the authors in \cite{cov29} concatenated Xception and ResNet50V2 and used 5-fold cross-validation techniques to detect COVID-19 virus. In another study, Ucar et al. \cite{cov30} used Bayes-SqueezeNet to develop a schema named COVIDiagnosis-Net to detect coronavirus using x-ray images. 
In another study \cite{cov32} the authors, presented a network called DarkCovidNet based on DarkNet backbone to detect COVID-19 virus using x-ray images. In another study, Punn et al. \cite{cov33} used ResNet, Inception-v3, Inception ResNet-v2, DenseNet-169, and NASNetLarge as a pre-trained CNN to detect COVID-19 virus from X-Ray images. While, Narin et al. \cite{cov34} used pre-trained models including  Inception-v3, ResNet50, Inception-ResNet-V2 to detect COVID-19 virus with 5-fold cross-validation in the dataset partition. The authors of another research work \cite{cov35} combined three pre-trained models such as ResNet-18, ResNet-50, and GoogleNet. While the authors in  \cite{cov38} used MobileNet, ResNet-50, VGG-16, VGG-19 backbones for COVID-19 detection method. In another study, \cite{cov39} the authors proposed a deep learning technique using ResNet50 to detect COVID-19.

\subsection{Panoptic segmentation}

Panoptic segmentation is a new direction in image segmentation also is the developed version of the instance and semantic segmentation. While the segmentation is make for the things and stuffs unlike instance segmentation that segment the things only. The panoptic segmentation models generate segmentation masks by holding the information from the backbone to the final density map without any explicit connections \cite{PS}. Many Backbones are used for images segmentation in general and for panoptic segmentation also, but the most used is ResNet family including ResNet-50 and ResNet-101.

Many methods have been proposed to segment the images using the panoptic presentation. From these works we can find the method in \cite{PS67} which is named fast panoptic segmentation network (FPSNet). Its a panoptic method while the instance segmentation and the merging heuristic part have been replaced with a CNN model called panoptic head. A feature map used to perform dense segmentation exploited ResNet-50 backbone.  Moreover, the authors in \cite{PS68} employ a position-sensitive attention layer which adds less computational cost instead of the panoptic head. It utilizes a stand-alone method based on the use of Deep-lab as backbone. While in \cite{PS69} a deep panoptic segmentation based on the bi-directional learning pipeline is utilized. Intrinsic interaction between semantic segmentation and instance segmentation is modeled using a bidirectional aggregation network called as BANet to perform a panoptic segmentation. The Backbone used here is ResNet-50. In the same context some authors worked on video instead of images. In  \cite{PS94}, video panoptic segmentation (VPSnet), which is a new video extension of panoptic segmentation is introduced, in which two types of video panoptic datasets have been used. The authors used as Backbone ResNet50 with FPN to extract feature map for the rest of network blocks. A holistic understanding of an image in the panoptic segmentation task can be achieved by modeling the correlation between object and background. For this purpose, a bidirectional graph reasoning network for panoptic segmentation (BGRNet) is proposed in \cite{PS95} Using ResNet-50 as well as the architecture in \cite{PS96} and \cite{PS97}. Using the same backbone in \cite{PS99}, the foreground things and background stuff have been dealt together in attention guided unified network (AUNet). Also using ResNet-50, an end-to-end occlusion-aware network (OANet) is introduced in \cite{PS100} to perform a panoptic segmentation, which uses a single network to predict instance and semantic segmentation. Without unifying the instance and semantic segmentation to get the panoptic segmentation, Hwang et al.\cite{PS102} exploited the blocks and pathways integration that allow unified feature maps that represent the final panoptic outcome. Finally, in \cite{PS107}, aiming at visualizing the hidden enemies in a scene, a panoptic segmentation method is proposed.
ResNet-101 is another Backbone used for panoptic segmentation in \cite{PS98} where the authors attempted to resolving overlaps using scene overlap graph network (SOGnet). In the same context, and using RestNet-101, a unified method named DR1Mask has been proposed in \cite{PS103} based on a shared feature map for both instance and semantic segmentation for panoptic segmentation.

While in \cite{PS106}, ResNeXt-101 for implementing the architecture of DetectoRS method.  DetectoRS is a panoptic segmentation method consists of two levels: macro level and micro level. At macro level, a recursive feature pyramid (RFP) is used to incorporate extra feedback connections from FPNs into the bottom-up backbone layers. While at the micro-level, a switchable atrous convolution (SAC) is exploited to convolve the features with different atrous rates and gather the results using switch functions. Using EfficientNet backbone, a variant of the Mask R-CNN as well as the KITTI panoptic dataset that has panoptic ground truth annotations are proposed in \cite{PS91}. The method named Efficient panoptic segmentation (EfficientPS). Also, a unified network named EffPS-b1bs4-RVC, which is a lightweight version of EfficientPS architecture is introduced in \cite{PS25}.

\subsection{Used backbones for each task}

In order to present the backbone used in each task, we attempted in this paper to collect the proposed methods and their feature extraction networks exploited. Table \ref{backbone-table} present the proposed methods for each task as well as the backbones used for each method. From the tables we can find that in some task some specific backbones are used widely for a task while are not exploited for others. For example, crowd counting methods are used VGG-16 network, while just some of them used VGG-19, MobileNet-v2, and ResNet-101. For Panoptic segmentation, ResNet-50 is the backbone that widely used fro segmenting the object in the panoptic way. GoogleNet and  ResNeXt are exploited for video summarization. For the others tasks, all the backbone are invested.

Figure \ref{figure:backbone-cerle} illustrate the computer vision tasks and the used backbones for each tasks. The used backbones have been presented by taking in account the number of paper used each backbone in a task. For example, for crowd counting, VGG is the most used backbone for counting the people or the object in monitored scene. The same observation for face recognition, while the almost proposed method exploit ResNet for extracting features. ResNet also the most used backbone for action recognition with GoogleNet. The same Backbone ResNet is the most used for panoptic segmentation and video summarization tasks. For object detection and COVID-19 detection tasks all the backbone are used equally unlike the other tasks that we can find a specific backbone widely used.

\section{Critical discussion} 

In order to present the impact of backbones on each task, some performance results are presented for image classification, object detection, action recognition, face recognition, panoptic segmentation, and video summarization. A comparison between the obtained results, is performed using the evaluation metrics used for each task as well as the datasets used for training and evaluation.
\begin{table}[t!]

\caption{Image classification performance results on ImageNet.}
\label{classeval}

\begin{tabular}{|l|p{2cm}|p{1cm}|p{1cm}|}
\hline

\textbf{Backbone} &	\centering\textbf{Complexity (GFLOPs)} &\centering \textbf{Top-1 err(\%)}&\textbf{\centering Top-5 err(\%)} \\
\hline
AlexNet \cite{AlexNet} &0.72& 42.8 & 19.7\\\hline
GoogLeNet &1.5& -&7.89\\\hline
VGG-16  \cite{VGG} & 15.3 &28.5& 9.9  \\\hline
ResNet-101 \cite{ResNet} &7.6 &19.87& 4.60\\\hline
ResNet-152 \cite{ResNet} & 11.3 &19.38 &4.49\\\hline
DenseNet-121 \cite{DenseNet} &0.525 & 25.02&  7.71 \\\hline
DenseNet-264 \cite{DenseNet} &1.125 &  22.15& 6.12  \\\hline
DetNet-50 \cite{Detnet} &3.8 &  24.1& -  \\\hline
DetNet-59 \cite{Detnet} &4.8 &  23.5& -  \\\hline
DetNet-101 \cite{Detnet} &7.6 &  23.0& -  \\\hline
SqueezeNet \cite{SqueezeNet}& 0.861 & 42.5 & 19.7 \\\hline
ResNeXt-50 \cite{ResNeXt}& 4.1 &22.2& - \\\hline
ResNeXt-101 (32x4d)\cite{ResNeXt}& 7.8&21.2& 5.6 \\\hline
ResNeXt-101 (64x4d)\cite{ResNeXt}& 15.6&20.4& - \\\hline
Xception \cite{Xception}& 11&21.0& 5.5 \\\hline
BN-Inception \cite{BNInception}&2&  22.0 &5.8\\\hline
Inception-v2 \cite{Inceptionv3}&-&  21.2 &5.6\\\hline
Inception-v3 \cite{Inceptionv3}&5.72&  \underline{18.7} &\underline{4.2}\\\hline
Inception-ResNet-v1 \cite{Inception-ResNet} &-&21.3 & 5.5\\\hline
Inception-v4 \cite{Inception-ResNet} &12.27& 20.0&5.0\\\hline
Inception-ResNet-v2 \cite{Inception-ResNet} &13.1& 19.9& 4.9\\\hline

WideResNet-50 \cite{WideResNet} &-&21.9 & 5.79\\\hline
MobileNet-224 \cite{Mobilenet}& 0.569 &29.4& -\\\hline
ShuffleNet-v1-50\cite{shuffleNet} &2.3 &25.2&-     \\\hline
ShuffleNet-v2-50 \cite{shuffleNetv2} & 2.3 &22.8&- \\\hline
EfficientNet-B0 \cite{efficientnet} & 0.39 &22.9& 6.7\\\hline
EfficientNet-B7 \cite{efficientnet} & 37 & \textbf{15.7}& \textbf{3.0} \\\hline
\end{tabular}

\end{table}

\subsection{Image classification evaluation} 

Image classification is the most task used for evaluating the new proposed CNN-based architecture. The most used dataset for this is ImageNet. The CNN-based architectures have been evaluated using generally three metrics including Top-1 error, Top-5 error, and the computational complexity of the model using the number of Flops. These metrics are used for comparing the effectiveness of the proposed models presented in Table \ref{classeval}. From the table we can see that some of the models evaluated just by Top-1 error and other just using Top-5 error. Also, the models that have high complexity reached good results. For example, EfficientNet-B7 \cite{efficientnet} have 37G for Flops which is generally high but reached minimum values of Top-1 and Top-5 error rates comparing with Inception-v3 \cite{Inceptionv3} that has the second best results with a difference of 3\%  in terms of Top-1 error rate and 1,2\% for Top-5 error rate. The same observation for the other models like Inception-ResNet-v2 \cite{Inception-ResNet}, ResNet-152 \cite{ResNet}, and  Inception-v4 \cite{Inception-ResNet}. For the low complexity methods we can find that the Top-1 and Top-5 error rates are higher than the others. For example, MobileNet has low complexity Folps, but the Top-1 error rate is high with value of 29.4, as well as AlexNet. Some of the method do not give the Flops number for their evaluation on ImageNet like WideResNet-50 \cite{WideResNet} , Inception-v2 \cite{Inceptionv3}, and Inception-ResNet-v1 \cite{Inception-ResNet}. From the Table \ref{classeval} and Figure \ref{fig:Appblck} we can find that the number of parameters and complexity of a model have an impact on the accuracy of it. Also, the development of different version of the same model increased the accuracy but with more complexity like in Inception-v3 \cite{Inceptionv3} and Inception-v4 \cite{Inception-ResNet}.

\begin{table*}[t!]
\begin{center}
\caption{Object detection performance results on MS COCO}
\label{table:objecteval}
\begin{tabular}{|p{0.41cm}|l|l|c|}
\hline
\textbf{Type} &\textbf{Method} &	\textbf{Backbone} &\textbf{mAP} \\
\hline
\multirow{8}{*}{\rotatebox[origin=c]{90}{Backbone}}& MobileNet-224 \cite{Mobilenet} & MobileNet &19.8\\\cline{2-4}
&ShuffleNet \cite{shuffleNet}  $2\times (g = 3)$  & ShuffleNet  & 25.0  \\\cline{2-4}
&Xception  \cite{Xception} & Xception &32.9 \\\cline{2-4}
&MobileNet-v2 \cite{Mobilenetv2} & MobileNet-v2 &  30.6  \\\cline{2-4}
&DetNet-50 \cite{Detnet} & DetNet& 37.9  \\\cline{2-4}
&DetNet-59 \cite{Detnet} & DetNet& \textbf{40.2} \\\cline{2-4}
&DetNet-101 \cite{Detnet} & DetNet& \underline{39.8}  \\\cline{2-4}
&ShuffleNet-v2 \cite{shuffleNetv2} & ShuffleNet-v2 &34.2 \\\cline{2-4}
&WideResNet-34 (tset) \cite{WideResNet}& WideResNet& 35.2 \\\hline\hline

\multirow{9}{*}{\rotatebox[origin=c]{90}{Backbone+Network}}&RetinaNet \cite{retinanet} &ResNet-101-FPN&39.1\\\cline{2-4}
&RetinaNet \cite{retinanet} &ResNeXt-101-FPN&40.8\\\cline{2-4}
&Faster R-CNN G-RMI \cite{23}&Inception-ResNet-v2 & 34.7\\\cline{2-4}
&Faster R-CNN TDM\cite{24}& Inception-ResNet-v2-TDM& 36.8 \\\cline{2-4}
&YOLOV2 \cite{27} & DarkNet-19 & 20.6\\\cline{2-4}
&YOLO-V3-spp \cite{yolov3} & DarkNet-19 & \textbf{60.6}\\\cline{2-4}
&YOLO-V4-P7 \cite{yolov4}&CSPDarknet53& \underline{55.7}\\\cline{2-4}
&EfficientDet-D7 \cite{EfficientDet}& EfficientNet-B7& \underline{55.1}\\\cline{2-4}

&DetectoRS \cite{Detectors}&ResNet-50  &53.0\\\cline{2-4}
&DetectoRS \cite{Detectors}& ResNeXt-101-64x4d  &55.7\\\hline
\end{tabular}
\end{center}
\end{table*}

\subsection{Object detection evaluation} 
The proposed deep learning methods for object detection are evaluated using mean Average Precision (mAP) metric on different datasets. The proposed DL-based methods used MS COCO for performance evaluation. This dataset have been taken into consideration due its wide usage for all the proposed object detection methods.

The evaluation of proposed methods on MS COCO dataset performed on val set using the mAP metric. For that, we attempted to present the obtained results for each method in Table \ref{table:objecteval}. In this table, we attempted to separate the methods into backbone methods and backbones+Network method. We mean by backbone methods, those pre-trained models that used object detection for proving the performance of the proposed networks then these methods are used also as backbone for other object detection methods. Backbones+Network denote the proposed methods that used famous backbones just for feature extraction step and they implement other blocks in their networks for detecting the objects. 
For the pre-trained models that evaluated on MS COCO for detecting the objects, we can see that DetNet-59 \cite{Detnet} achieved the best results with a mAP of 40.2 followed by  DetNet-101 \cite{Detnet} by difference of 0.4. While MobileNet-v1 \cite{Mobilenet} is the lowest reached results of a mAP of 19.8. The others method including ShuffleNet \cite{shuffleNet}, Xception  \cite{Xception}, MobileNet-v2 \cite{Mobilenet}, and ShuffleNet-v2 \cite{shuffleNetv2} reached close results.

For the backbone+Network methods, we can find from the table that YOLO-V3-spp \cite{yolov3}, YOLO-V4-P7 \cite{yolov4}, , and DetectoRS \cite{Detectors} reached the first best results, while YOLO family used DarkNet as backbone and DetectoRS used ResNet and ResNeXt as backbone. The results using EfficientDet-D7 \cite{EfficientDet} are also good with a difference of 0.6 compared with DetectoRS and YOLO-v3.

According to mAP values achieved, we can find that the performance of these methods using DL still need improvement even the number of works proposed. This is due to the complexity of scenes that can contain many objects with different situation like small scales, different positions, occlusion between objects, etc.

\subsection{Face recognition evaluation}
In order to evaluate the proposed face recognition methods, many datasets are used. The most used datasets includes LFW and YTF that are common face recognition datasets. For that, we collected some of DL and DRL-based methods tested on the same datasets by mentioning the backbones used as feature extraction models. Table ~\ref{faceeval} represent some obtained results by face recognition  approaches on LFW and YTF datasets. From the table we can observe that the performance accuracies are close for DL and DRL-based approaches. For example, the DL-based methods  ArcFace \cite{fra18} and FairLoss \cite{fra19} comes in the first place by an accuracies of 99.83 and 99.75 on LFW and 98.02 and 96.2 for YTF respectively. The others methods are also in the same range, while we can find the difference between the accuracies not exceed 0.5\% for LFW and 5\% for YTF.  The same observation for DRL-based methods that reached more than 99\% on LFW and 96\% for YTF.

Another observation for DL and DRL-based approaches, the same backbone family is used for almost all methods while the depth of the networks is varying. For these backbones we can find ResNet-20, ResNet-50 and ResNet-100. While VGG is used by Range loss \cite{fra13}.   
\begin{table*}[t!]
\caption{{Face verification accuracy (\%) on LFW  and YTF datasets trained on WebFace dataset.} }
\label{faceeval}
\def\arraystretch{1}
\ignorespaces 
\centering 
\begin{tabular}{|l|l|l|l|l|}
\hline
\textbf{Type}&\textbf{Method} & \textbf{Backbone} & \textbf{LFW} & \textbf{YTF} \\ \hline

\multirow{3}{*}{DL}&CosFace(LMCL) \cite{fra12} (2018)	& ConvNet& 99.33	& 96.1  \\\cline{2-5}
&Range loss \cite{fra13}	& VGG-19 & 99.52	& 93.7  \\\cline{2-5}
&NRA+CD \cite{fra15} &ResNet-50  & 99.53 &	96.04 \\\cline{2-5}
&RegularFace \cite{fra17}	& ResNet-20 & 99.33 &	94.4  \\\cline{2-5}
&ArcFace \cite{fra18} &ResNet-100  & \textbf{99.83} & \textbf{98.02}  \\\cline{2-5}
&FairLoss  \cite{fra19}  &ResNet-50  & \underline{99.57} & \underline{96.2}  \\\cline{1-5}\hline\hline
\multirow{2}{*}{DRL}&Attention-aware-DRL \cite{fra20} & ResNet&- & \textbf{96.52} \\\cline{2-5}
&Margin-aware-DRL \cite{fra21} & ResNet-50&\textbf{99.57} & \underline{96.2}  \\\cline{2-5}\hline
\end{tabular}\par 
\end{table*}


\subsection{Action recognition evaluation }

Action recognition methods based on Deep learning (DL) and Deep reinforcement learning (DRL) are presented in \ref{actioneval}. These methods are collected based on the backbones and datasets used. For example, we choose for comparison the dataset used by more than two DL and DRL-based methods, because some datasets are used in just one paper. The results presented in this comparison are on UCF-101 and HMDB51 datasets.  In order to evaluate their methods the performance accuracy is invested by the proposed architectures on UCF-101 and HMDB51.

The obtained results on each dataset are presented in Table ~\ref{actioneval}. On HMDB-51 dataset, the DL-based method in \cite{ar06} that exploited ResNeXt-101 as backbone reached the highest accuracy,  while \cite{ar02} is the second best result by an accuracy of 73.3. The same method \cite{ar02} achieved the best performance accuracy on UCF-101, while \cite{ar06} results comes in the second place by a difference of 1.9 point. For \cite{ar06} they used ResNet-50 as backbone. For I3D \cite{ar06} method, ConvNet is used as backbone while the obtained results is less than \cite{ar06} by 8.6 point for on HMDB-51 dataset and 2.4 points for UCF-101 dataset.

\begin{table*}[t!]
\caption{{Performance of action recognition methods on HMDB-51  and UCF-101 datasets} }
\label{actioneval}
\def\arraystretch{1}
\ignorespaces 
\centering 
\begin{tabular}{|l|l|l|l|l|}
\hline
\textbf{Type} & \textbf{Method} & \textbf{Backbone} & \textbf{HMDB-51} & \textbf{UCF-101} \\ \hline

\multirow{3}{*}{\rotatebox[origin=c]{90}{DL}}&TEA  \cite{ar02}& ResNet-50  & \underline{73.3} & \textbf{96.9} \\\cline{2-5}
&I3D \cite{ar05}&3D ConvNet & 66.4 &93.4  \\\cline{2-5}

&Li et al.  \cite{ar06}& ResNeXt-101 &\textbf{75.0} & \underline{95.0}\\\cline{1-5}\hline\hline

\multirow{4}{*}{\rotatebox[origin=c]{90}{DRL}}&PEAR  \cite{ar2}&  BN-Inception  & - & 84.99 \\\cline{2-5}

&PA-DRL \cite{ar3} &  VGG-16 &- &87.7 \\\cline{2-5}

&Li et al.  \cite{ar4}& GoogleNet &\underline{66.8} & \underline{93.2}\\\cline{2-5}

&TSN-AS \cite{ar5}& BN-Inception & \textbf{71.2} &  \textbf{94.6}\\\cline{2-5}

&Wang et al.  \cite{ar6}&GoogleNet &60.6 &- \\\cline{2-5}
\cline{1-3}
\end{tabular}\par 
\end{table*}

\begin{table*}[t!]
\caption{Performance comparison of existing panoptic segmentation schemes on the val set under Cityscapes and COCO datasets.}
\label{table:cc}
\begin{center}

\begin{tabular}{|p{1cm}|p{5.cm}|p{.7cm}|p{.7cm}|p{.7cm}|p{.7cm}|p{.7cm}|p{.7cm}|p{.7cm}|p{.7cm}|p{.7cm}|}
\hline
\multirow{2}{*}{\textbf{Dataset}} &
 \multirow{2}{*}{\textbf{Method/Backbone}} &		\multicolumn{3}{|c|}{\textbf{PQ}} & \multicolumn{3}{|c|}{\textbf{SQ}}&	\multicolumn{3}{|c|}{\textbf{RQ}}			\\ \cline{3-11}

& &\textbf{PQ} & \textbf{PQ$^{st}$} & \textbf{PQ$^{th}$} & \textbf{SQ} & \textbf{SQ$^{st}$} & \textbf{SQ$^{th}$} & \textbf{RQ} & \textbf{RQ$^{st}$} & \textbf{RQ$^{th}$} \\ 

\hline
\multirow{8}{*}{Cityscapes}& FPSNet \cite{PS67} / ResNet-50  &		55.1&	60.1&	48.3& -&-&-&-&-&- \\\cline{2-11}

&Axial-DeepLab \cite{PS68} / DeepLab  & \textbf{67.7}  & -& -&-  & - & - & -& - & - \\\cline{2-11}

& EfficientPS Single-scale \cite{PS91} / EfficientNet &	63.9& 66.2	&\underline{60.7}	& \underline{81.5}& \underline{81.8} & \underline{81.2} & \underline{77.1} &\underline{79.2}&\underline{74.1} \\\cline{2-11}
	
&EfficientPS Multi-scale \cite{PS91} / EfficientNet &	\underline{65.1}&  \textbf{67.7}	& 	 \textbf{61.5}	& \textbf{82.2} & 	 \textbf{82.8}& 	 \textbf{81.4} &  	 \textbf{79.0} & 	 \textbf{81.7}& 	 \textbf{75.4}  \\\cline{2-11}

&VPSNet \cite{PS94} / ResNet-50 &  62.2 & 65.3& 58.0 &-  & - & - & - & - & - \\\cline{2-11}
 
&SpatialFlow \cite{PS96} / ResNet-50 &58.6 &  61.4& 54.9 &-  & - & - & - & - & - \\\cline{2-11}  

& SPINet \cite{PS102} / ResNet-50  &63.0&  \underline{67.3}  &57.0 & -& - & - & - & - & - \\\cline{2-11}   

&Son et al. \cite{PS107} / ResNet-50 &58.0&  - &- & 79.4& - & - & 71.4 & - & - \\\hline\hline


\multirow{9}{*}{COCO} & Axial-DeepLab \cite{PS68} / DeepLab  & \underline{43.9} & \textbf{36.8} &48.6 &-  & - & - & -& - & - \\\cline{2-11}

&BANet \cite{PS69} / ResNet-50 & 	43.0& 31.8	&50.5	 &	\underline{79.0} &\textbf{75.9}&\underline{81.1}& \underline{52.8}& \textbf{39.4}& \textbf{61.5} \\\cline{2-11}

&	BGRNet \cite{PS95} / ResNet-50	&43.2 & 33.4& 49.8 &-&-&-&-&-&- \\\cline{2-11}

 & SpatialFlow \cite{PS96} / ResNet-50 &40.9 &  31.9& 46.8 &-  & - & - & - & - & - \\\cline{2-11}
 
 & Weber et al. \cite{PS97} / ResNet-50&	32.4&	28.6&	34.8& - &-&-&-&-&-\\\cline{2-11}
 
 &SOGNet \cite{PS98} / ResNet-102 &	43.7&	33.2&\underline{50.6}& -&-&-&-&-&- \\\cline{2-11}
 
& OANet \cite{PS100} / ResNet-50&	40.7&	26.6&	50.0&	78.2&	\underline{72.5}&		\textbf{82.0}&	49.6&	\underline{34.5}&	\underline{59.7} \\\cline{2-11}

& SPINet \cite{PS102} / ResNet-50  &42.2&  31.4  &49.3  & -& - & - & - & - & -\\\cline{2-11}  

& DR1Mask \cite{PS103} / ResNet-101  & \textbf{46.1}&  \underline{35.5} &\textbf{53.1}    &  \textbf{81.5}& - & - &    \textbf{55.3}  & - & - \\\hline 

\end{tabular}
\end{center}
\end{table*}
For DRL-based methods, the performance evaluations are performed on the same two datasets and presented in ~\ref{actioneval}. For example, TSN-As \cite{ar5} is the highest accuracy on HMDB-51 followed by \cite{ar4} with a difference of 4.4 points using BN-Inception and GoogleNet respectively. The same observation on UCF-101, while the same methods \cite{ar5} and \cite{ar4} achieved the highest accuracy values of 94.6 and 93.2 respectively.

From the table, We can observe that the DL and DRL-based action recognition methods tested on different datasets using different backbones, have a difference in terms of performance accuracies reached. While DL-based method achieved better results than DRL-based methods. Also for DL-based methods we found that the method used ResNet and ResNeXt reached higher accuracies than the other used GoogleNet or VGG. which means the use specific backbones for a specific task can make the difference in terms of performance.

\subsection{Panoptic segmentation evaluation}

Cityscapes and COCO datasets are the most commonly preferred datasets for experimenting the efficiency of panoptic segmentation solutions. A detailed report on the methods that use thess datasets with the evaluation metrics are given in Table \ref{table:cc}. In addition, the obtained results have been presented considering the backbones used. Though it is common to use the val set for reporting the results. Here we presented just the evaluation on val set. All the models are representative, and the results listed in Table \ref{table:cc} have been published in the reference documents.

On Cityscapes, we can observe that different methods, such as \cite{PS107} and \cite{PS103} have evaluated their results using the three metrics, i.e. PQ, SQ, and RQ. While some approaches have also been evaluated using these metrics on things ($PQ^{th}$, $SQ^{th}$, and $RQ^{th}$) and stuff ($PQ^{st}$, $SQ^{st}$, and $RQ^{st}$), e.g. EfficientPS \cite{PS91}. Moreover, from the results in Table \ref{table:cc}, Axial-DeepLab \cite{PS68} reaches the highest PQ values, with an improvement of 2.6\% than the second-best result obtained by EfficientPS Multi-scale \cite{PS91}. Regarding the SQ metric, EFFIcientPS achieves the best result using single-scale and multi-scale,  by diffrence of 0.7\%. The same thing Using SQ metric, EfficientPS provides the best accuracy results. The difference between EfficientPS and other methods is that it utilizes a pre-trained model on Vistas dataset, where the schemes do not use any pre-training. In addition, EfficientPS uses EfficientNet as backbone while the most of discussed techgniques have exploited RestNet-50 except Axial-Deeplab, which uses DeepLab as a backbone.

On the COCO val set, the evaluation results are slightly different from the obtained results on Cityscapes dataset although there are some frameworks that have reached the highest performance, such as DR1Mask \cite{PS103} for PQ, $PQ^{st}$, and SQ, Axial-DeepLab \cite{PS68} for $PQ^{st}$, BANet \cite{PS69} for $SQ^{st}$, $RQ^{st}$, and $SQ^{th}$, OANet \cite{PS100} for $SQ^{th}$. For example, using DR1Mask, the performance rate for using PQ and PQ$^{th}$ metrics has reached 46.1\% and 53.1\%, respectively. The difference between the methods that reaches the highest results and those in the second and third places, is around 1-4\%, which demonstrates the effectiveness of these panoptic segmentation schemes.

\begin{table*}[t!]
\caption{{The performance of each method on the existing crowd counting dataset. The \textbf{bold} and \underline{underline} fonts respectively represent the \textbf{first} and \underline{second} place } }
\def\arraystretch{1.2}
\label{crowdeval}
\footnotesize
\ignorespaces 
\centering 
\begin{tabular}{|l|c|cc|cc|cc|cc|}
\hline

 && \multicolumn{2}{c|}{\textbf{ShanTech\_A }} & \multicolumn{2}{c|}{\textbf{ShanTech\_B }}& \multicolumn{2}{c|}{\textbf{UCF\_QNRF}}& \multicolumn{2}{c|}{\textbf{UCF\_CC\_50}} \\

\cline{3-4} \cline{5-6}\cline{6-10}
\textbf{Method} &\textbf{Backbone} &\textbf{MAE} & \textbf{MSE} & \textbf{MAE} & \textbf{MSE}& \textbf{MAE} & \textbf{MSE}& \textbf{MAE} & \textbf{MSE}\\

\hline
CSRNet \cite{count1} (2018) & VGG-16 & 68.2 & 115.0 & 10.6 & 16.0 & - & - & 266.1 & 397.5	\\ \hline
SPN  \cite{count2} (2019)	&VGG-16	 &	61.7 &  99.5	 & 9.4  & 14.4	 & -& 	-& 	259.2 &	335.9 \\ \hline
DENet\cite{count4} (2020)	&VGG-16	&  65.5 &  101.2 &  9.6 & 15.4& - & -&  241.9 &  345.4	\\ \hline
CANNet \cite{count5}(2019)	&VGG-16	& 62.3 & 100.0 & 7.8 &\underline{12.2}	&  107.0 & 183.0&  212.2 & 243.7\\
\hline
SCAR \cite{count6} (2019)  &VGG-16& 66.3 & 114.1& 9.5 & 15.2 & -&-& 	259.0 & 374.0 \\ \hline

ADNet  \cite{count21} (2020) & VGG-16 & 61.3 &103.9 & 7.6 &12.1 & 90.1 & \underline{147.1} & 245.4 &327.3\\ \hline

ADSCNet  \cite{count21} (2020) &  VGG-16 & \textbf{55.4}& \underline{97.7} & \textbf{6.4} & \textbf{11.3} & \textbf{71.3} &\textbf{132.5} & 198.4& 267.3\\ \hline

ASNet  \cite{count22} (2020) & VGG-16 & \underline{57.7} & \textbf{90.1} &- &- & 91.5& 159.7 & \textbf{174.8} &\underline{251.6} \\ \hline

SCNet  \cite{count9} (2021)& VGG-16&58.5&99.1&8.5&13.4&93.9&150.8&\underline{197.0}&\textbf{231.6}\\ \hline

BL  \cite{count7} (2019)&VGG-19 & 62.8 &101.8 & \underline{7.7} &12.7 & \underline{88.7} &154.8  & 229.3 &308.2\\ \hline

MobileCount \cite{count3} (2020)& MobileNet-V2 &  84.8 & 135.1  	  & 8.6  & 13.8		 & 127.7 &  216.5	 & 284.5 & 421.2 		\\
\hline
SFCN  \cite{count8} (2019)& ResNet-101 & 64.8& 107.5& 7.6 &13.0 &102.0 &171.4 & 214.2 &318.2\\ \hline

\hline
\end{tabular}
\end{table*}

\subsection{Crowd counting evaluation}

The obtained results using MAE and MSE metrics are presented in Table \ref{crowdeval}. From this table we can observe that many methods succeed to estimate the number of people in the crowd with promising results especially for ShanTech\_Part\_B dataset due to simple crowd density in this dataset and all the images contains the same depth of the crowd and the same distribution of the people in the scenes. We can find also that the ShanTech\_Part\_A comes in the second place in terms of MAE reached due to the same reasons of ShanTech\_Part\_B but here the images are more crowded that the images in ShanTech\_Part\_B. For the other datasets including UCF\_QNRF, and UCF\_CC\_50, the images are more crowded which can reach 4000 people per images with make the estimation of the density maps more complex. Also the the scale and shape variations in these dataset affect the performance of each method. 

For the obtained MAE and MSE of each method, the results in the table shows that each method reach good results in a dataset better than the others. And this comes from the treatment used for each method as well as the backbone exploited for feature extraction. For example some method are working on the scale-variation while others used segmentation of the crowd region before estimating the crowd density. For the ADSCNet method, we can see that it outperform the others method in three dataset including  ShanTech\_Part\_A, ShanTech\_Part\_B, and UCF\_QNRF with an MAE of 55.4 on ShanTech\_Part\_A and less by 2.3 point than the ASNet method which come in the second place. While we can find that the SPN, SCAR, CANNet method reached close results of MAE values. On UCF\_CC\_50 dataset, SCNet achieved the less MSE results better of ASNet with 20 point. For these results we can conclude that the methods used VGG-16 are most effective methods comparing with the method used MobileNet and ResNet-101 as backbones. Also, some proposed architectures can be better in some cases while its not in others like ADSCNet on hanTech\_Part\_A, ShanTech\_Part\_B, and UCF\_QNRF datasets and AsNet and SCNet on UCF\_CC\_50 dataset.   

\begin{table*}[t!]
\begin{center}
\caption{Comparison of the performance of video summarization methods}
\label{vsummeval}
\begin{tabular}{|l|l|c|c|c|}
\hline
\textbf{Type}&\textbf{Method} &\textbf{Backbone} &\textbf{SumMe} &\textbf{TVSum}  \\
\hline
\multirow{3}{*}{\rotatebox[origin=c]{90}{DL}} &GoogleNet+Transf \cite{vs1}&  GoogleNet& \underline{51.6} & \underline{64.2}\\\cline{2-5}
&ResNet+Transf \cite{vs1}&  ResNet&\textbf{ 52.8 } & \textbf{65.0}  \\\cline{2-5}

&MCSF \cite{vs2}&GoogleNet& 48.1& 56.4\\\cline{2-5}
&Zhang et al. \cite{vs10} & ResNet-152 &  37.7 &  51.1  \\\cline{1-5}\hline\hline

\multirow{3}{*}{\rotatebox[origin=c]{90}{DRL}} &Lei et al. \cite{vs7}& AlextNet& 41.2 & 51.3  \\\cline{2-5}

&Chen et al. \cite{vs8}& GoogleNet&\textbf{43.6} & \underline{58.4}  \\\cline{2-5}

&DR-DSN \cite{vs11}& GoogleNet  &41.4  & 57.6 \\\cline{2-5}
&DR-DSN-s \cite{vs11}& GoogleNet  &42.1  & 58.1 \\\cline{2-5}
&Wang et al. \cite{vs12}&GoogleNet & \underline{43.4}& \textbf{58.5} \\\cline{2-5}
&SGSN \cite{vs13}  & Inception-V3  &41.5 &  55.7  \\ \hline

\end{tabular}
\end{center}
\end{table*}

\subsection{Video summarization evaluation} 

To show the performance of each video summarization method using DL and DRL, The obtained results using state-of-the-art methods on SumMe and TVSum datasets are presented is Table \ref{vsummeval}. From the table, we can find that GoogleNet and ResNet are the most used feature extraction backbones  For deep learning and deep reinforcement learning based approaches. For DL-based approaches, the method in \cite{vs1} achieved the best results on SumMe and TVSum datasets. While, ResNet+Transformer \cite{vs1} comes in the fisrt place by a values of 52.8\% and 65.0\% on SumME and TVSum, and  GoogleNet+Transformer \cite{vs1} reached the second best results by a deference of 1.2\% for the two datasets. For the other approaches like MCSF \cite{vs2} and  \cite{vs10} the obtained results achieved close results with a difference of 9.6\% for SumMe and 5.3\% for TVSum. Generally all the method are in the same range in term of accuracy reached on the two dataset.

For DRL-based methods including \cite{vs7}, \cite{vs8}, DR-DSN \cite{vs11}, DR-DSN-s \cite{vs11}, \cite{vs12}, and SGSN \cite{vs13}, the obtained result are close while we cane find the difference between the Best  and worst result on SumMe dataset not exceed 2.4\% and 7.1\% for TVSum. On SumMe dataset, the method in \cite{vs8} reached 43.6\% which is the best result, followed by \cite{vs12} with a value of 43.4\%. We can observe that the wto method used GoogleNet as backbone. The same methods reached the best results on TVSum but this time \cite{vs12} reached the first best results and \cite{vs8} comes in the second place with 58.5\%.

From the presented results we can find that the performance of these methods using DL and DRL is challenging according to the accuracy rate achieved. In addition, the performance of these methods on TVSum is more improved than SumMe dataset.


\subsection{COVID-19 detection evaluation}
To  demonstrate  the  performance  of  proposed COVID-19 detection methods , many metrics have been used including accuracy, specificity, and sensitivity. Some method used transfer learning using many pre-trained models. We collect a set of these method to compare the impact of each model on COVID-19 detection dataset. Table \ref{covid} represent a set of method providing the obtained results with the three metrics. From the table we can find that all the methods succeed to detection COVID-19 from X-ray images with convinced accuracies. While \cite{cov34} using Inception-v3 reached the highest results of 100\% for sensitivity and specificity metrics and 99.5\% for model accuracy. Using SqueezeNet in \cite{cov30} the obtained results comes in the second places with a difference from 1-2\% comparing with  \cite{cov34} that used Inception-v3. We can find also that, the used model have different results even using the same architectures, due to the representation of data as well as the pre-possessing operations performed before start the training. Unlike the other tasks like crowd counting and panoptic segmentation, COVID-19 proposed methods do not have any special backbone used for detection and all the backbone have been used. 

\section{Challenges and future directions}

\begin{table*}
\begin{center}
\caption{COVID-19 detection techniques using different method that based on cited backbones. }
\label{covid}
\begin{tabular}{|l|c|c|c|c|c|c|}
\hline
\textbf{Authors}& \textbf{Model} &\textbf{Classes} &		\textbf{Accuracy} &	\textbf{Sensitivity} &	\textbf{Specificity}\\

\hline
& MobileNet-v2& 3 &96.22 &96.22 & 97.80 \\ \cline{2-6}
Chowdhury et al.\cite{cov24}& Inception-v3& 3 &96.2 & 96.4 & 97.5 \\ \cline{2-6}
& ResNet-101&3&96.22 & 96.2 & 97.8 \\ \cline{2-6}
& DenseNet-201& 3 &97.9 & 97.9 & 98.8 \\ \hline

Ucar et al. \cite{cov30} & SqueezeNet& 3 & \underline{98.3} &\underline{98.3} & 99.1\\ \hline
Ozturk et al. \cite{cov32}& DarkNet& 3	& 98.1 & 95.1 & 95.3 \\  \hline

&Inception-v2 &3 & 88.0 & 79.0 & 89.0 \\ \cline{2-6}
Punn et al. \cite{cov33}&Inception-ResNet-v2 &3 &92.0 & 92.0 & 89.0 \\ \cline{2-6}
&DenseNet-169 & 3 & 95.0 & 96.0 & 95.0  \\\hline

& Inception-v3&3	&\textbf{99.5} & \textbf{100} & \textbf{100} \\ \cline{2-6}
& ResNet-50 &3&91.7 &57.0 &91.3 \\ \cline{2-6}
Narin et al. \cite{cov34} & ResNet-152  & 3 & 97.3 &  93.2 &\underline{99.3} \\ \cline{2-6}
& Inception-ResNet-v2 & 3&96.3 & 78.0 & 96.8 \\ \hline

Ozcan et al. \cite{cov35}& ResNet-50&  4 & 97.6 & 97.2 & 97.9 \\ \hline

\end{tabular}
\end{center}
\end{table*}

\subsection{Deep learning challenges}

Deep learning is a trending technology for all computer science and robotics tasks to help and assist human actions. Using artificial neural networks, that suppose to work like a human brain, deep learning is an aspect of AI that consists of solving the classification and recognition goals for making machines learn from specific data for specific scenarios. Deep learning has many challenges even the development reached in different tasks. For that, a list of deep learning challenges will be discussed in this section. 

\textbf{Size of data used for learning}:
A large-scale dataset is a necessary condition for a deep learning model to work well. Also, the performance of such a deep learning system is related to the size of data used. For that, the annotation and availability of data are real challenges for deep learning methods. For example, we can find many tasks while the data can not be available for the public like for industrial applications, or a task has few scenarios, also for medical purposes some times the data size is small due to the uniqueness of some diseases.

\textbf{Non-contextual Understanding:}
The capability of  a deep learning model is related to the architecture used which is deep and contains many layers and levels, but it is not related to the level of understanding of it. For example, if a model is proficient in a specific task, and to use this trained model in another close task, all the training and the processing should be re-trained because this model does not understand the context, but lean what it is trained on only. Also, with the development in different domains, a deep learning model should be maintained every time with the new features and data to understand the new scenarios.

\textbf{Data labeling and annotation:}
In CV, the segmentation of scenes and objects in an image or video represents a crucial challenge. For automatic
segmentation, data should be prepared first by annotating
and labeling the object or the scenes of interest before starting the training of such a method. The annotations represent also a challenge, due to the number of objects that should be labeled, also any changes to the scenes require another labeling according to the types of the objects and the categories of scenes \cite{f1,f2,f22}.

\subsection{Future directions}
Nowadays, DL and DRL techniques are used not just for analyzing the content of images but also replacing the work of the human, like making the decisions and annotating the data. in this section, a couple of future directions of DL and DRL is discussed.   

\textbf{Data augmentation}:

The lack of large-scale datasets for some tasks represents a challenge for deep learning and deep reinforcement learning methods. For that, the researchers started using DL and DRL techniques for data augmentation especially for medical imaging that suffer from the lack of data for many diseases. For example,  DRL is used for creating new images to be used for training like in \cite{f3}. While the authors proposed a DRL architecture for kidney Tumor segmentation. The proposed method starts by augmenting the data before using it for segmentation.

\textbf{Data annotations:}
Image and video annotation is a big challenge for researchers specialty for computer vision tasks which need an enormous effort for annotating the objects, labeling the scenes and object for segmentation, or separating the classes for image classification. Also, the format of the annotations can be different from a type of method to another. For example, We can find that for object detection many methods like DetectronV2, YOLO, or EfficientDet used several formats including txt, XML, DarkNet, or JSON formats. A common format for all the methods can be used to overcome this problem.  Also, in order to automatize the annotations process for object segmentation in a video sequence, the authors in \cite{f4} proposed a DRL method using an extended version of Dueling DQN. The researchers also started labeling the data for image segmentation using DRL-based techniques. The obtained results using this method are not effective at this time due to the complexity of the purpose, as well as it is the first method that attempted to label the video and
images instead of using manual labeling by humans. But with the development of the DRL technique in different computer vision applications, automatic data labeling and annotations can be reached.

\section{Conclusion}
This paper presents an overview of deep learning networks used as a backbone for many proposed architectures for computer vision tasks. A detailed description is provided for each network. In addition, some computer vision tasks are discussed regarding the backbone used for extracting the features. We attempted also to collect the experimental results for each method within each task and comparing them based on the backbone used. This review can help the researcher because it a detailed summarization and comparison of the famous backbones also link to the code-sources are given. In addition, a set of DL and DRL challenges are presented with some future directions.

\section*{Acknowledgments}
This publication was made possible by NPRP grant \#
NPRP12S-0312-190332 from Qatar National Research Fund
(a member of Qatar Foundation). The statement made herein
are solely the responsibility of the authors.







\end{document}